\documentclass{article} %
\usepackage{iclr2027_conference,times}

\usepackage[utf8]{inputenc}
\usepackage[T1]{fontenc}
\usepackage{hyperref}
\hypersetup{hidelinks}
\usepackage{url}
\usepackage{booktabs}
\usepackage{graphicx}
\usepackage{amsmath}
\usepackage{amssymb}
\usepackage{amsfonts}
\usepackage{microtype}
\usepackage{xcolor}
\usepackage{multirow}
\usepackage{xspace}
\usepackage{caption}
\usepackage{subcaption}
\usepackage{enumitem}
\usepackage[table]{xcolor}
\usepackage{array}
\usepackage{tcolorbox}
\tcbuselibrary{listings}
\usepackage{makecell}
\usepackage{placeins}
\usepackage{float}
\usepackage{flafter}
\usepackage{needspace}
\usepackage{adjustbox}

\newcolumntype{C}[1]{>{\centering\arraybackslash}p{#1}}
\newcolumntype{L}[1]{>{\raggedright\arraybackslash}p{#1}}
\newcolumntype{R}[1]{>{\raggedleft\arraybackslash}p{#1}}

\definecolor{holisticBlue}{HTML}{C9DAF8}
\definecolor{temporalGreen}{HTML}{D9EAD3}
\definecolor{entityYellow}{HTML}{FFF2CC}
\definecolor{comparativePeach}{HTML}{E4D7F5}

\colorlet{holisticBlueFrame}{holisticBlue!70!black}
\colorlet{temporalGreenFrame}{temporalGreen!70!black}
\colorlet{entityYellowFrame}{entityYellow!75!black}
\colorlet{comparativePeachFrame}{comparativePeach!70!black}

\newcommand{\st}{Spatial Tracking\xspace}
\newcommand{\ta}{Temporal Alignment\xspace}
\newcommand{\cor}{Comparative Reasoning\xspace}
\newcommand{\hs}{Holistic Synthesis}

\newcommand{\newtext}[1]{#1}

\DeclareCaptionFont{redfont}{\color{black}}
\DeclareCaptionFont{appendixblackfont}{\color{black}}

\newcommand{\stcS}{\colorbox{entityYellow}{Spat. Tracking\xspace}}
\newcommand{\tacS}{\colorbox{temporalGreen}{Temp. Alignment\xspace}}
\newcommand{\corcS}{\colorbox{comparativePeach}{Comp. Reasoning\xspace}}
\newcommand{\hscS}{\colorbox{holisticBlue}{Holi. Synthesis\xspace}}

\newcommand{\taskCount}{Object Counting}
\newcommand{\taskRoute}{Route Planning}
\newcommand{\taskOrder}{Sequential Ordering}
\newcommand{\taskSync}{Multi-Angle Synchronization}
\newcommand{\taskReID}{Object Re-identification}
\newcommand{\taskMeasure}{Spatial Measurement}
\newcommand{\taskNumerical}{Numerical Comparison}
\newcommand{\taskKinematic}{Kinematic Comparison}

\newcommand{\taskcCount}{\colorbox{holisticBlue}{Object Counting}}
\newcommand{\taskcRoute}{\colorbox{holisticBlue}{Route Planning}}
\newcommand{\taskcOrder}{\colorbox{temporalGreen}{Sequential Ordering}}
\newcommand{\taskcSync}{\colorbox{temporalGreen}{Multi-Angle Synchronization}}
\newcommand{\taskcReID}{\colorbox{entityYellow}{Object Re-identification}}
\newcommand{\taskcMeasure}{\colorbox{entityYellow}{Spatial Measurement}}
\newcommand{\taskcNumerical}{\colorbox{comparativePeach}{Numerical Comparison}}
\newcommand{\taskcKinematic}{\colorbox{comparativePeach}{Kinematic Comparison}}

\newtcblisting{promptbox}[1][]{
  colback=gray!10,
  colframe=gray!80,
  listing only,
  listing options={
    basicstyle=\ttfamily\small,
    breaklines=true,
    breakatwhitespace=true
  },
  title={#1},
  boxrule=0.5pt,
  arc=2mm,
  width=\linewidth,
  left=1mm, right=1mm,
  top=1mm, bottom=1mm
}

\tcbset{
  pillar style/.style 2 args={
    colback=#1!45,
    colframe=#2,
    listing only,
    listing options={
      basicstyle=\ttfamily\small,
      breaklines=true,
      breakatwhitespace=true
    },
    boxrule=0.5pt,
    arc=2mm,
    width=\linewidth,
    left=1mm, right=1mm,
    top=1mm, bottom=1mm
  }
}

\newtcblisting{temporalbox}[1][]{pillar style={temporalGreen}{temporalGreenFrame}, title={#1}}
\newtcblisting{spatialbox}[1][]{pillar style={entityYellow}{entityYellowFrame}, title={#1}}
\newtcblisting{comparativebox}[1][]{pillar style={comparativePeach}{comparativePeachFrame}, title={#1}}
\newtcblisting{holisticbox}[1][]{pillar style={holisticBlue}{holisticBlueFrame}, title={#1}}

\title{SYNCR: Diagnosing and Learning Cross-Video Reasoning from Simulation}

\author{Sara Ghazanfari, Siddharth Garg, Prashanth Krishnamurthy, Farshad Khorrami \\
New York University \\
\texttt{sg7457@nyu.edu}
}

\iclrfinalcopy
\begin{document}

\maketitle
\lhead{Preprint}

\vspace{-4mm}
\begin{abstract}
Reasoning across videos requires aligning events, matching identities, comparing motion, and integrating partial observations. Evaluating these capabilities and testing how to improve them requires both reliable labels and targeted supervision. We introduce SYNCR, a simulator-grounded framework that connects these two needs through shared task generators. Built on Habitat, Kubric, and CLEVRER, SYNCR derives answers from environment state and provides 4,000 evaluation questions and 15,960 training questions over disjoint videos, spanning eight cross-video reasoning tasks. Visual ablations and human evaluation assess dependence on the supplied evidence and answer recoverability. 
Evaluation of 22 multimodal large language models reveals persistent difficulties in physical comparison and scene integration that increasing model size does not consistently resolve. Supervised fine-tuning raises Qwen3-VL-8B’s average SYNCR accuracy from 32.6\% to 61.6\%, with gains extending to task configurations and video sources absent from training for those tasks. Transfer to real footage is most consistent for temporal ordering: accuracy improves by 9.0–20.5 percentage points on constructed Assembly101 and Panoptic ordering sets across three checkpoints spanning two model families and two model sizes, with additional gains on existing temporal reasoning benchmarks. These results establish SYNCR as a controlled setting for diagnosing cross-video reasoning failures, testing their learnability, and identifying where synthetic supervision transfers.

\end{abstract}
\vspace{-3mm}
\section{Introduction}
\label{sec:intro}

Multiple videos can provide complementary evidence about an environment, but understanding each video independently is insufficient to answer questions about their relationships. Counting distinct objects across overlapping camera views requires recognizing repeated sightings; reconstructing an event from shuffled clips requires relating its progression across segments. Such tasks demand cross-video reasoning: aligning events, matching identities, comparing behavior, and integrating partial observations. Which of these capabilities remain difficult for current multimodal large language models, and can targeted supervision improve them?

Answering these questions requires connecting evaluation with controlled learning experiments. Real-footage benchmarks such as MVU-Eval, CVBench, and CrossVid capture diverse multi-video settings, but precise temporal offsets, three-dimensional relationships, and kinematic quantities are difficult to annotate at scale. Ambiguous visual evidence and textual shortcuts can also complicate the interpretation of model errors and successes. Simulation offers access to the state underlying each observation, enabling both programmatic labeling and the generation of targeted supervision. However, a simulator-derived answer is useful for evaluating visual reasoning only if it is recoverable from the supplied videos. A complementary framework must therefore combine grounded labels, evidence controls, and tests of generalization beyond its training conditions.

We introduce SYNCR, a simulator-grounded framework that uses the same task generators to diagnose cross-video reasoning failures and test their learnability. Built on Habitat, Kubric, and CLEVRER, SYNCR derives answers from temporal offsets, object identities, trajectories, and scene geometry, while models receive only RGB videos, questions, and answer options. Eight tasks cover temporal alignment, spatial tracking, comparative reasoning, and holistic synthesis. The generators produce 4,000 evaluation questions and 15,960 training questions over disjoint videos. Text-only and single-video ablations assess shortcuts and dependence on multiple inputs, while human evaluation assesses whether answers are recoverable from the RGB videos.

Across 22 models, SYNCR reveals uneven capabilities: chronological ordering can be strong while physical comparison and scene integration remain difficult, including at larger model sizes. These failures are nevertheless responsive to targeted supervision. On matched sets of 200 items per task, eight-task supervised fine-tuning raises Qwen3-VL-8B’s average accuracy from 32.6\% to 61.6\%, and improves performance under changes to task structure and video source. The framework thus connects observed weaknesses to direct tests of whether training can address them.

Real-video evaluations then test the reach of these improvements. Temporal ordering transfers across constructed Assembly101 and Panoptic sets, with gains of 9.0–20.5 percentage points across three checkpoints from two model families. Improvements on existing temporal reasoning benchmarks provide further evidence beyond our constructed evaluations, while results on other tasks remain mixed. Together, these experiments contribute grounded task generators, matched evaluation and training resources, and a controlled study of which cross-video capabilities can be improved through synthetic supervision and where those improvements generalize.
\begin{figure}[t]
  \centering
  \includegraphics[width=0.90\linewidth,trim={0 3pt 0 28pt},clip]{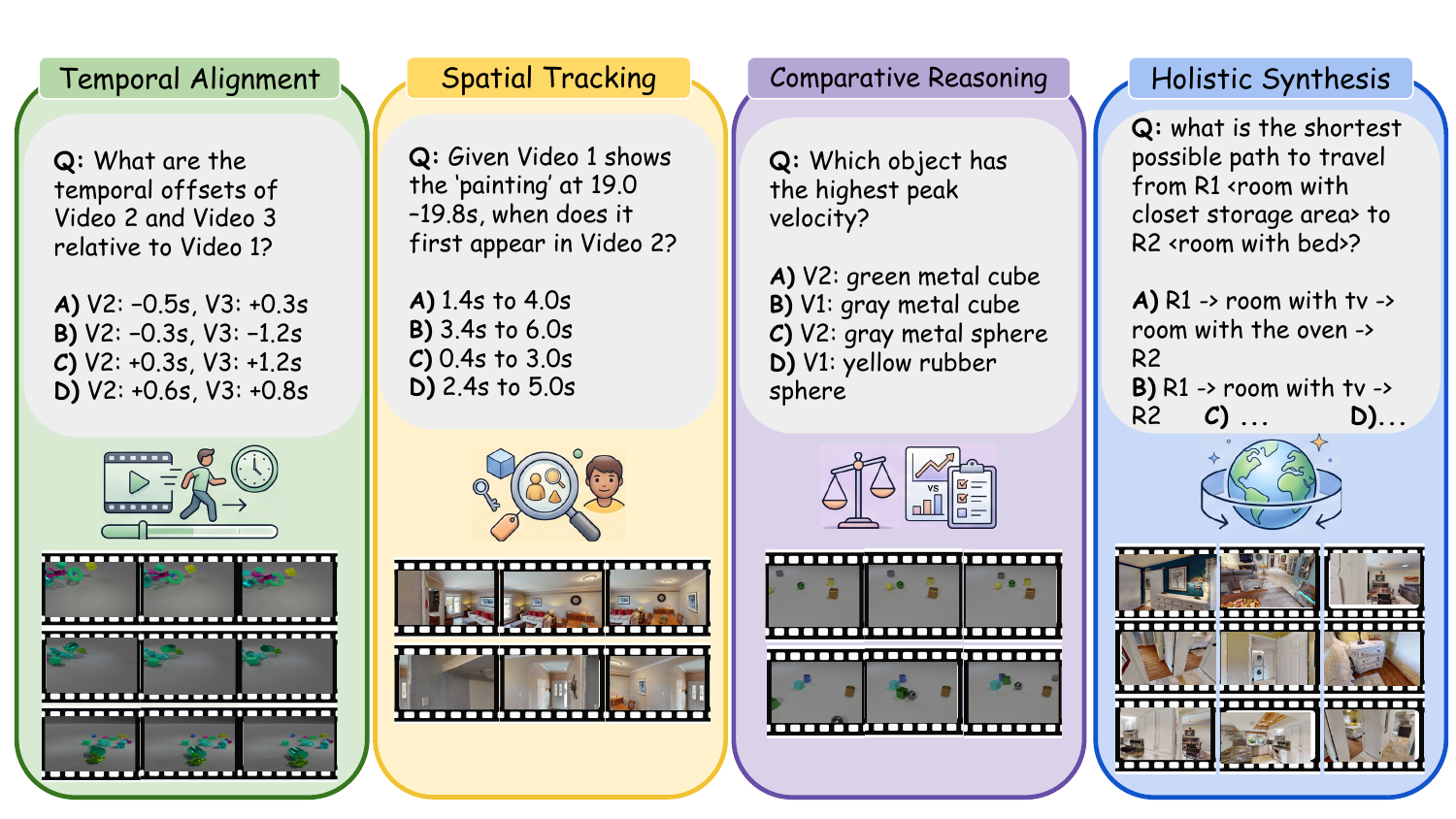}
  \caption{\textbf{The SYNCR framework.} Eight tasks in four families: temporal alignment across streams, identity and geometry across viewpoints, physical and numerical comparison, and integration of partial scene observations. Simulator-derived answers support both evaluation and targeted training.}
  \label{fig:syncr-teaser}
\end{figure}
\vspace{-3mm}
\section{SYNCR: Tasks and Data Construction}
\label{sec:framework}

SYNCR uses eight task generators for both evaluation and training. Each generator renders or selects multiple video clips, computes an answer from simulator state, and constructs a question designed to require relating the clips. This shared construction lets us measure a task-level failure and then test whether supervision on that task improves it. Evaluation and training use disjoint videos; later experiments vary task structure and video source to test whether gains extend beyond the training.

\subsection{Diagnostic Task Families}
\label{sec:pillars}
We group the eight tasks into four diagnostic families (Fig.~\ref{fig:syncr-teaser}). \ta relates events across time; \st connects identities and positions across views; \cor compares motion or events between videos; and \hs{} combines partial observations of a scene. The families organize the task-level results; they are not assumed to be independent abilities. This organization exposes failures that an eight-task average would hide.

\textbf{\ta.} \textit{\taskSync} presents three independently cropped camera views of one event and asks for the start times of Videos~2 and~3 relative to Video~1. Solving it requires matching moments despite changes in viewpoint. \textit{\taskOrder} instead shuffles four segments from one continuous video and asks for their chronological order. Gaps between segments prevent matching adjacent boundary frames directly; the model must use event progression and object motion. Ordering therefore tests a different temporal relation from synchronization across cameras.

\textbf{\st.} \textit{\taskReID} presents two indoor walkthroughs, identifies an object's visibility window in Video~1, and asks when the same object first appears in Video~2. \textit{\taskMeasure} shows a dynamic scene from two cameras and asks which object is closest in 3D to a target when another object leaves a designated view. The departure fixes the time of comparison; the two projections must be reconciled because image-plane proximity need not match 3D distance.

\textbf{\cor.} \textit{\taskKinematic} asks which of four candidate objects---the two fastest in each of two videos---reaches the highest peak speed. \textit{\taskNumerical} asks which video has more collisions and by how many, allowing ties. Both compare evidence across videos, but one depends on estimating motion over time and the other on counting discrete events.

\textbf{\hs.} \textit{\taskCount} asks for the number of distinct objects in two or three named categories across three indoor walkthroughs; repeated sightings of one object count once. \textit{\taskRoute} presents three partial route clips and asks for the shortest path between rooms identified by their objects. Counting requires linking repeated sightings, while routing requires joining room connections shown in separate clips.
\vspace{-3mm}
\subsection{Simulator-Grounded Item Construction}
\label{sec:data}

Each generator uses simulator annotations to select clips, compute the answer, and build answer options. Shared instance IDs establish correspondence across views; positions, velocities, and collision records determine spatial and physical answers; and crop start times determine temporal labels. Models receive only RGB clips, questions, and options. We use the annotations to label and filter items, not as model input. Exact simulator labels do not by themselves establish that an answer is visible, so Sec.~\ref{sec:sanity} and App.~\ref{app:dataset_details} report evidence controls and implementation details.

\textbf{Habitat.} We render 100-frame egocentric trajectories at 5~FPS in the 216 densely annotated HM3D scenes~\citep{ramakrishnan2021habitat}; each trajectory ends with a panoramic sweep. Per-frame instance masks supply the visibility windows for \taskReID\ and the distinct instance counts for \taskCount, while navigation graphs supply \taskRoute\ answers. Rooms are described by their three rarest objects. We require re-identified objects to appear for at least five consecutive frames and counted objects to occupy at least 5\% of a frame in one view, filtering out instances with little visual exposure.

\textbf{Kubric.} We simulate 10--15 moving objects with collisions and render each scene for 5~s at 12~FPS from three synchronized cameras. Independent 3-second crops set the \taskSync\ offsets relative to Video~1. The cameras share object identities, but visibility is computed separately for each view. For \taskMeasure, simulator positions give 3D distances at the time of the event specified in the question.

\textbf{CLEVRER.} We reuse the original videos and derive answers from their trajectory, velocity, and collision annotations. \taskOrder\ uses four shuffled clips with 0.3--0.6~s gaps. For \taskKinematic, we require a speed margin of at least 0.25 between the cross-video maximum and its competitors and between each video's two fastest objects, reducing near ties. \taskNumerical\ compares annotated collision counts; ties make up one fifth of its evaluation items.

\textbf{Distractor Design.} We construct incorrect options to resemble valid answers: synchronization distractors come from independently sampled crop offsets, re-identification distractors from visibility windows with similar timings, and route distractors from paths with similar hop counts. Ordering alternatives follow the answer permutation frequencies; numerical alternatives use incorrect video identities and count differences. These choices limit cues in the options, which we assess with text-only audits (App.~\ref{app:blind}).

\subsection{Evaluation and Training Splits}
\label{sec:stats}
The benchmark contains 500 questions per task: 4,000 questions referencing 4,827 distinct video files across tasks. The training set contains 15,960 questions referencing 14,956 distinct video files across tasks, with no video file shared between the evaluation and training splits (Table~\ref{tab:dataset_stats}; App.~\ref{app:dataset_stats}). Each question presents two to four clips. Answer categories and option letters are balanced both in the first 200 and across all 500 evaluation items per task, limiting answer-frequency cues. The held-out videos test learning beyond the rendered training examples.

\begin{table*}[!t]
     \centering
    \caption{\textbf{Zero-shot accuracy (\%) on SYNCR.} Avg is the unweighted mean across eight tasks. The 17 standard open-weight checkpoints and the Thinking checkpoint use 500 questions per task; proprietary models$^\dagger$ and the four-person human majority vote use the same first 100. Bold and underline mark the best and second-best results among the 17 standard open-weight checkpoints. Chance is 25\%, except for \textsc{Num} (20\%).}

  \label{tab:main_results}
  \resizebox{1.0\textwidth}{!}{
    \begin{tabular}{L{34mm}*{8}{C{10mm}}C{6mm}}
    \toprule
    \multirow{2}{*}{\textbf{Model}} & \multicolumn{2}{c}{\textbf{\tacS}} & \multicolumn{2}{c}{\textbf{\stcS}} & \multicolumn{2}{c}{\textbf{\corcS}} & \multicolumn{2}{c}{\textbf{\hscS}} & \multirow{2}{*}{\textbf{Avg}} \\
    \cmidrule(lr){2-3} \cmidrule(lr){4-5} \cmidrule(lr){6-7} \cmidrule(lr){8-9}
    & {\footnotesize\textsc{Sync}} & {\footnotesize\textsc{Order}} & {\footnotesize\textsc{ReID}} & {\footnotesize\textsc{Meas}}  & {\footnotesize\textsc{Num}} & {\footnotesize\textsc{Kin}} & {\footnotesize\textsc{Count}} & {\footnotesize\textsc{Route}} & \\
    \midrule
Human & 100.0 & 100.0 & 92.0 & 76.0 & 100.0 & 68.0 & 92.0 & 88.0 & 89.5 \\

Gemini-3-Flash$^\dagger$ & 28.0 & 84.0 & 72.0 & 32.0 & 24.0 & 4.0 & 28.0 & 48.0 & 40.0 \\
Gemini-3.1-Pro$^\dagger$ & 60.0 & 96.0 & 68.0 & 48.0 & 16.0 & 8.0 & 32.0 & 40.0 & 46.0 \\
GPT-5.4$^\dagger$ & 36.0 & 64.0 & 48.0 & 36.0 & 20.0 & 16.0 & 32.0 & 44.0 & 37.0 \\
GPT-6 Astra$^\dagger$ & 88.0 & 100.0 & 88.0 & 52.0 & 60.0 & 20.0 & 48.0 & 60.0 & 64.5 \\
    \midrule
    \multicolumn{10}{l}{\textbf{Model Size $\leq$ 4B}} \\
    \midrule
LLaVA-OV 0.5B          & 25.8 & 24.4 & 26.4 & 23.0 & 16.0 & 24.0 & 25.4 & 25.4 & 23.8 \\
InternVL3.5 1B         & 24.4 & 25.8 & 24.6 & 24.4 & 20.0 & 27.6 & 22.8 & 27.4 & 24.6 \\
InternVL3.5 2B         & 26.8 & 25.2 & 26.0 & 31.2 & 26.8 & \underline{29.6} & 28.4 & 28.0 & 27.8 \\
Qwen3-VL 2B            & 24.2 & 28.4 & 33.6 & 29.4 & 21.4 & 20.2 & 25.6 & 23.2 & 25.8 \\
InternVL3.5 4B         & 26.4 & 30.6 & 26.2 & 24.8 & 29.2 & 24.2 & 23.6 & 25.6 & 26.3 \\
Qwen3-VL 4B            & 26.2 & 49.4 & 46.4 & \underline{38.0} & 26.6 & 20.0 & 28.8 & 28.0 & 32.9 \\
    \midrule
    \multicolumn{10}{l}{\textbf{4B $<$ Model Size $\leq$ 8B}} \\
    \midrule
LLaVA-OV 7B            & 27.2 & 21.6 & 26.6 & 27.6 & 20.8 & 24.2 & 26.0 & 23.6 & 24.7 \\
Qwen2.5-VL 7B          & 25.0 & 36.8 & 25.8 & 29.8 & 22.4 & 21.2 & 27.2 & 25.6 & 26.7 \\
InternVL3 8B           & 26.0 & 35.4 & 25.0 & 29.6 & 26.6 & 26.0 & 27.2 & 20.0 & 27.0 \\
InternVL3.5 8B         & 26.4 & 39.4 & 26.0 & 31.4 & 31.6 & \textbf{29.8} & 26.4 & 24.0 & 29.4 \\
Qwen3-VL 8B            & 29.6 & 56.4 & \underline{47.6} & 32.6 & 29.6 & 17.2 & \underline{30.2} & 29.4 & 34.1 \\
Qwen3-VL 8B Thinking & 23.8 & 76.6 & 47.0 & 24.0 & 23.0 & 20.6 & 28.6 & 32.0 & 34.4 \\
    \midrule
    \multicolumn{10}{l}{\textbf{Model Size $>$ 8B}} \\
    \midrule
InternVL3.5 14B        & \textbf{31.0} & 43.6 & 31.2 & 36.6 & 32.2 & 27.0 & 25.8 & 25.0 & 31.6 \\
Qwen2.5-VL 32B         & 27.6 & 30.2 & 32.8 & 30.2 & 23.6 & 21.8 & 28.8 & 31.2 & 28.3 \\
Qwen3-VL 32B           & \underline{30.0} & \underline{64.4} & \textbf{56.2} & 35.2 & \underline{33.6} & 19.4 & 28.6 & \underline{32.4} & \textbf{37.5} \\
InternVL3.5 38B        & 26.0 & 44.4 & 34.6 & \textbf{39.0} & \textbf{41.0} & 26.2 & 30.0 & \textbf{32.8} & \underline{34.3} \\
LLaVA-OV 72B           & 21.8 & 26.0 & 31.2 & 30.0 & 22.6 & 18.6 & 27.4 & 23.6 & 25.2 \\
Qwen2.5-VL 72B         & 27.4 & \textbf{66.8} & 31.8 & 31.6 & 33.0 & 24.4 & \textbf{30.4} & 25.8 & 33.9 \\
    \bottomrule
    \end{tabular}%
  }
\end{table*}

\section{Evaluating Cross-Video Reasoning}
\label{sec:experiments}

We first ask which SYNCR tasks current models solve, then test whether their answers depend on the intended video evidence. Table~\ref{tab:main_results} reports the task-level results; Fig.~\ref{fig:validity_checks} summarizes the visual-evidence controls.

\subsection{Experimental Setup}
\label{sec:experimental_setup}

\textbf{Models.} The 22 evaluated models comprise 17 standard open-weight checkpoints (0.5B--72B parameters) from Qwen2.5-VL~\citep{Qwen2.5-VL}, Qwen3-VL~\citep{bai2025qwen3}, InternVL3~\citep{zhu2025internvl3}, InternVL3.5~\citep{wang2025internvl3}, and LLaVA-OneVision~\citep{li2024llava}; Qwen3-VL-8B-Thinking; and four proprietary models: Gemini-3-Flash~\citep{gemini3flash2025}, Gemini-3.1-Pro~\citep{gemini31pro2026}, GPT-5.4~\citep{openai2026gpt54}, and GPT-6~Astra~\citep{openai2026gpt6astra}.

\textbf{Protocol.} Models answer zero-shot multiple-choice questions with inputs labeled \textit{Video~1}, \textit{Video~2}, etc. We score the parsed final answer and use deterministic decoding where supported (App.~\ref{app:evaluation_details}). The 17 standard open-weight checkpoints and Qwen3-VL-8B-Thinking receive 500 items per task. Proprietary models and the human reference receive the same first 100. Chance is 25\% on seven tasks and 20\% on \taskNumerical. We compare models on matched items when testing differences and report 95\% Wilson intervals for accuracy and exact two-sided McNemar tests for paired comparisons (App.~\ref{app:zeroshot}).

\textbf{Human reference.} Four graduate students independently answer the 100-item sets from RGB videos, questions, and options, without simulator annotations. We report majority-vote accuracy and score tied votes as incorrect (App.~\ref{app:human_baseline}). This four-person reference measures what viewers can recover together; it is not an estimate of individual human accuracy.

\subsection{Main Results}
\label{sec:main_results}

\textbf{Large gaps remain on the matched human items.} GPT-6~Astra averages 64.5\%, versus 89.5\% for the human majority vote on the same 100 items per task (Table~\ref{tab:main_results}). Among the standard open-weight checkpoints, Qwen3-VL~32B has the highest eight-task average, 37.5\% on 500 items per task; seven of the 17 average below 27\%. The 100-item and 500-item scores use different subsets and should not be treated as a precise cross-group ranking.

\textbf{The two temporal tasks separate sharply.} On \taskOrder, GPT-6~Astra reaches 100\% and Qwen3-VL-8B-Thinking 76.6\%. On \taskSync, GPT-6~Astra reaches 88\% and Gemini-3.1-Pro 60\%, but the 17 standard open-weight checkpoints score only 21.8--31.0\%, around the 25\% chance level. Ordering gapped segments from one recording and matching moments across camera views therefore present different difficulties.

\textbf{Physical comparison and scene integration remain difficult.} The highest standard open-weight score on \taskKinematic\ is 29.8\%; no proprietary model exceeds 20\% on its 100-item set. The human majority vote reaches 68\%, suggesting that this task is difficult even for unaided viewers. On \taskCount\ and \taskRoute, standard open-weight scores reach at most 30.4\% and 32.8\%, while the human majority vote reaches 92\% and 88\% on its matched set. These task-level gaps identify where aggregate accuracy hides persistent failures.

\noindent\begin{minipage}[t]{0.53\linewidth}
\vspace{0pt}
\textbf{The Thinking checkpoint has mixed temporal results.} Against Qwen3-VL-8B-Instruct on the same 500 items per task, the Thinking checkpoint raises \taskOrder\ from 56.4\% to 76.6\% but lowers \taskSync\ from 29.6\% to 23.8\% (App.~\ref{app:thinking}). Because the checkpoints differ in post-training, this comparison does not isolate the effect of deliberation alone.

\textbf{Scaling gains vary by model family.} Qwen3-VL's average rises from 25.8\% at 2B to 37.5\% at 32B, and InternVL3.5's from 24.6\% at 1B to 34.3\% at 38B; LLaVA-OneVision stays between 23.8\% and 25.2\% across 0.5B--72B (Fig.~\ref{fig:avg_scaling}). Within Qwen3-VL, the clearest gains are on \taskOrder\ and \taskReID, while several other tasks change little or non-monotonically.
At Qwen3-VL sizes of 4B, 8B, and 32B, \taskKinematic\ remains below chance, while \taskMeasure\ and \taskCount\ vary non-monotonically (App.~\ref{app:zeroshot}). Model size alone therefore does not remove every task-level failure.
\end{minipage}\hfill
\begin{minipage}[t]{0.44\linewidth}
\vspace{0pt}
\centering
\includegraphics[width=0.92\linewidth]{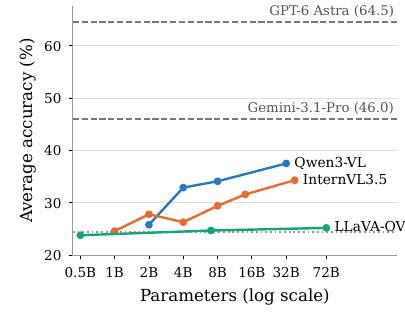}
\captionof{figure}{\textbf{Zero-shot SYNCR accuracy by model size.} Points show the eight-task mean for each plotted open-weight checkpoint (500 items per task). Dashed lines mark GPT-6~Astra and Gemini-3.1-Pro (100 items per task); the dotted line marks the 24.4\% mean chance rate.}
\label{fig:avg_scaling}
\end{minipage}
\subsection{Validity Checks}
\label{sec:sanity}

\begin{figure}[!htbp]
    \centering
    \includegraphics[width=\linewidth]{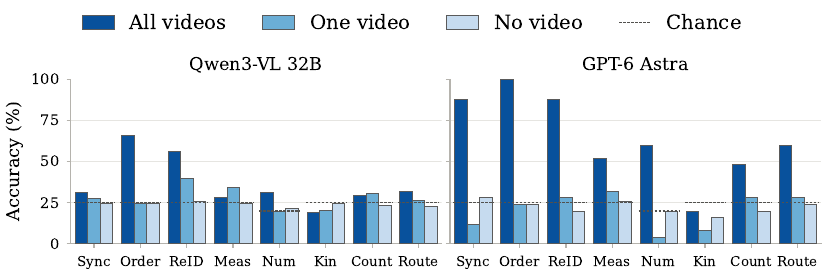}
    \caption{\textbf{Visual-evidence controls (zero-shot).} For each item, the question and options stay fixed while the model receives all videos, one randomly selected video, or no video. Qwen3-VL-32B's all-video and one-video results use the first 200 items, while its no-video result uses all 500; the latter is therefore not a paired comparison.}
    \label{fig:validity_checks}
\end{figure}

Simulator state supplies exact labels, but the evaluation must also test whether a model needs the intended visual evidence. We remove all videos or retain one randomly selected video while holding the question and options fixed; we also check whether crop timestamps reveal answers (Fig.~\ref{fig:validity_checks}; App.~\ref{app:sanity}).

\textbf{Text alone provides little signal.} Without videos, Qwen3-VL 8B and 32B and LLaVA-OneVision~72B remain within the chance interval on every task. GPT-6~Astra, explicitly asked to guess, averages 22.3\% and exceeds the chance interval on no task. Audited text-only rules also fail to solve the tasks reliably (App.~\ref{app:blind}). These zero-shot controls do not establish that trained models remain free of textual shortcuts; Sec.~\ref{sec:training_controls} tests that separately.

\textbf{Removing clips reduces accuracy on several tasks.} Keeping one randomly selected video reduces Qwen3-VL~32B's \taskOrder\ accuracy from 66.0\% to 24.5\%, and GPT-6~Astra's \taskSync\ accuracy from 88\% to 12\% and \taskNumerical\ accuracy from 60\% to 4\%. The reduction is not universal: Qwen3-VL~32B still reaches 39.5\% on \taskReID\ with one video, versus 56.0\% with all videos. For tasks already near chance with all videos, this ablation has little ability to reveal a further loss. The controls support a multi-video interpretation most clearly where full-input accuracy is above chance.

\textbf{Source timestamps do not reveal temporal labels.} Training and evaluation videos are disjoint. Source-video timestamps are removed from cropped inputs, while elapsed time within each clip is retained. Thus, timestamp markers cannot directly reveal the \taskOrder\ sequence or \taskSync\ offsets (App.~\ref{app:timestamp_leak}).

\newtext{\textbf{SYNCR rankings correlate with real-video scores.} To test whether the synthetic benchmark tracks performance beyond its own items, we evaluate the same 17 standard open-weight checkpoints on MVU-Eval Temporal Reasoning and Comparison using the same harness. Their SYNCR averages correlate with both real-video scores (Spearman $\rho=0.86$ and $0.92$, both $p<0.001$); the correlations remain $0.78$ and $0.87$ after accounting for model size (App.~\ref{app:construct_validity}). This supports SYNCR's relevance to real-video evaluation. A general video-understanding factor could also produce these correlations, so they do not establish that SYNCR isolates cross-video reasoning.}

\section{Learning from Synthetic Supervision}
\label{sec:learning}

The same task generators used for diagnosis provide supervision. We test three increasingly demanding outcomes: improvement on held-out SYNCR videos, generalization to changed task structures and video sources, and transfer to real footage.

\begin{figure}[t]
  \centering
  \includegraphics[width=1.0\linewidth]{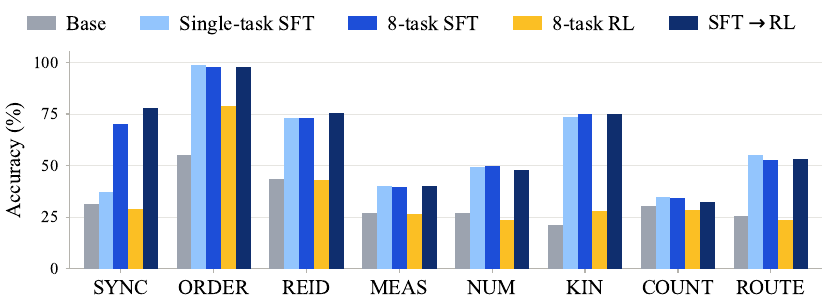}
  \caption{\textbf{SYNCR accuracy after finetuning Qwen3-VL-8B.} All bars use the same 200 held-out items per task. Each single-task SFT bar comes from a separate adapter trained on that task; eight-task SFT uses one adapter trained on all tasks. Eight-task RL applies GRPO to the base checkpoint, while SFT$\rightarrow$RL applies it after eight-task SFT.}
  \label{fig:training_bars}
\end{figure}

\textbf{Training setup.}
\label{sec:training_setup}
We train Qwen3-VL-8B-Instruct for one epoch with rank-16 LoRA adapters~\citep{hu2022lora} and a frozen vision encoder (App.~\ref{app:training_details}). Eight single-task adapters each receive one task's examples; an 8-task adapter receives all 15,960 training examples. We repeat the 8-task recipe on Qwen3-VL-4B and InternVL3.5-8B (App.~\ref{app:other_models}). We also test GRPO~\citep{shao2024deepseekmath} with binary correctness rewards, initialized either from the base model or from 8-task SFT (App.~\ref{app:other_models}).

\subsection{Training Gains on SYNCR}
\label{sec:learnable}

\textbf{Supervision improves seven of eight tasks.} On the same first 200 held-out items per task for base and trained models, Qwen3-VL-8B rises from 32.6\% to 61.6\% average accuracy after 8-task SFT (Fig.~\ref{fig:training_bars}). It reaches 98\% on \taskOrder, 75\% on \taskKinematic, 73\% on \taskReID, and 70\% on \taskSync. Seven task gains survive correction for multiple comparisons (App.~\ref{app:multiplicity}). \taskCount\ is the exception: it changes from 30.5\% to 34.5\%, a smaller gain than on the other tasks. Removing the answer options reveals better relative count estimates despite weak exact accuracy (App.~\ref{app:count_openended}).

\textbf{Synchronization is higher with multi-task SFT.} \taskSync\ reaches 70.0\% with eight tasks and 69.5--78.0\% in other mixes that include it, versus 37.0\% with \taskSync-only SFT; larger training sets preclude attributing the difference to task combination alone.
The 8-task recipe raises InternVL3.5-8B from 29.3\% to 62.8\% on matched items. Qwen3-VL-4B reaches 62.6\% on the first 200 items per task, but its available base score uses all 500, so a matched gain is unavailable (App.~\ref{app:other_models}).

\textbf{The tested GRPO recipe yields smaller gains.} Eight-task GRPO from the base improves the mean by 2.5 points, or 4.2 with a second seed, versus 28.9 points for 8-task SFT. Single-task GRPO raises \taskKinematic\ from 21.0\% to 48.5\%, still below the corresponding SFT adapter's 73.5\% (App.~\ref{app:single_task_rl}). Starting GRPO from 8-task SFT reaches 62.5\%, 0.9 points above SFT alone. This comparison concerns the tested binary-reward GRPO configuration, not RL methods in general.

\subsection{Generalization to New Structures and Video Sources}
\label{sec:transfer_within}
\begin{table}[t]
  \centering
  \color{black}
  \caption{\textbf{Generalization to shifted task configurations.} Accuracy (\%) of the base and eight-task SFT Qwen3-VL-8B. ``Standard'' uses the first 200 benchmark items; ``New'' uses a separate 200-item set. (a) Changes the number of segments, videos, or queried categories, or uses a held-out camera rig for \textsc{Sync}. (b) Tests a task on a simulator absent from that task's training examples (C: CLEVRER; K: Kubric; H: Habitat).}
  \label{tab:shift_generalization}
  \footnotesize
  \setlength{\tabcolsep}{3pt}
  \begin{minipage}[t]{0.43\linewidth}
    \centering
    \textit{(a) Structure and camera-rig shifts}\\[2pt]
    \begin{tabular}{lcccc}
    \toprule
    & \multicolumn{2}{c}{\textbf{Standard}} & \multicolumn{2}{c}{\textbf{New}} \\
    \cmidrule(lr){2-3}\cmidrule(lr){4-5}
    \textbf{Task}, train $\rightarrow$ test & \textbf{Base} & \textbf{SFT} & \textbf{Base} & \textbf{SFT} \\
    \midrule
    \textsc{Order} 4 $\rightarrow$ 3 & 55.0 & 98.0 & 64.0 & 99.0 \\
    \textsc{Order} 4 $\rightarrow$ 5 & 55.0 & 98.0 & 64.0 & 99.5 \\
    \textsc{Sync} 3 $\rightarrow$ 2 & 31.5 & 70.0 & 26.5 & 56.0 \\
    \textsc{Sync} rig A $\rightarrow$ B & 31.5 & 70.0 & 23.0 & 74.5 \\
    \textsc{Num} 2 $\rightarrow$ 3 & 27.0 & 50.0 & 23.5 & 45.5 \\
    \textsc{Kin} 2 $\rightarrow$ 3 & 21.0 & 75.0 & 12.5 & 91.0 \\
    \textsc{Count} 2--3 $\rightarrow$ 4 & 30.5 & 34.5 & 32.0 & 36.0 \\
    \bottomrule
    \end{tabular}
  \end{minipage}\hfill
  \begin{minipage}[t]{0.55\linewidth}
    \centering
    \textit{(b) Held-out task--simulator pairings}\\[2pt]
    \begin{tabular}{lcccc}
    \toprule
    & \multicolumn{2}{c}{\textbf{Standard}} & \multicolumn{2}{c}{\textbf{New}} \\
    \cmidrule(lr){2-3}\cmidrule(lr){4-5}
    \textbf{Task}, train $\rightarrow$ test & \textbf{Base} & \textbf{SFT} & \textbf{Base} & \textbf{SFT} \\
    \midrule
    \textsc{Order} C $\rightarrow$ K & 55.0 & 98.0 & 35.5 & 95.5 \\
    \textsc{Order} C $\rightarrow$ H (walk) & 55.0 & 98.0 & 41.5 & 50.5 \\
    \textsc{Order} C $\rightarrow$ H (route) & 55.0 & 98.0 & 55.5 & 71.0 \\
    \textsc{Count} H $\rightarrow$ K (shape) & 30.5 & 34.5 & 24.5 & 40.0 \\
    \textsc{Count} H $\rightarrow$ C (shape) & 30.5 & 34.5 & 56.5 & 57.0 \\
    \textsc{Count} H $\rightarrow$ C (mat.+shape) & 30.5 & 34.5 & 49.0 & 58.0 \\
    \bottomrule
    \end{tabular}
  \end{minipage}
\end{table}

We evaluate fixed 200-item sets with either a changed task structure or a new video source for that task (Table~\ref{tab:shift_generalization}; App.~\ref{app:shift}). Structural shifts change the number of segments, videos, or queried categories while retaining the simulator and question format. A separate \taskSync\ shift uses a held-out camera rig with altered positions and focal lengths. Source swaps keep the task format but use an engine absent from that task's training examples.

\textbf{Gains persist under structural and camera-rig shifts.} With three or five segments instead of four, trained \taskOrder\ accuracy reaches 99.0--99.5\%. With three videos instead of two, \taskKinematic\ rises from 12.5\% to 91.0\% and \taskNumerical\ from 23.5\% to 45.5\%. Two-video \taskSync\ rises from 26.5\% to 56.0\%; on the held-out camera rig it reaches 74.5\% versus 23.0\% for the base. \taskCount\ shows a smaller change on the four-category shift.

\textbf{Transfer to held-out task--simulator pairings varies.} Ordering improves from 35.5\% to 95.5\% on Kubric, but more modestly on Habitat walkthroughs (41.5\% to 50.5\%) and route clips (55.5\% to 71.0\%). InternVL3.5-8B likewise rises from 55.0\% to 100.0\% on Kubric ordering (App.~\ref{app:shift}). \taskCount\ improves on Kubric shape counting (24.5\% to 40.0\%) and CLEVRER material--shape counting (49.0\% to 58.0\%), but changes little on CLEVRER shape counting.

\subsection{Sim-to-Real Transfer}
\label{sec:sim2real_transfer}

\begin{figure}[h]
    \centering
    \includegraphics[width=0.9\linewidth]{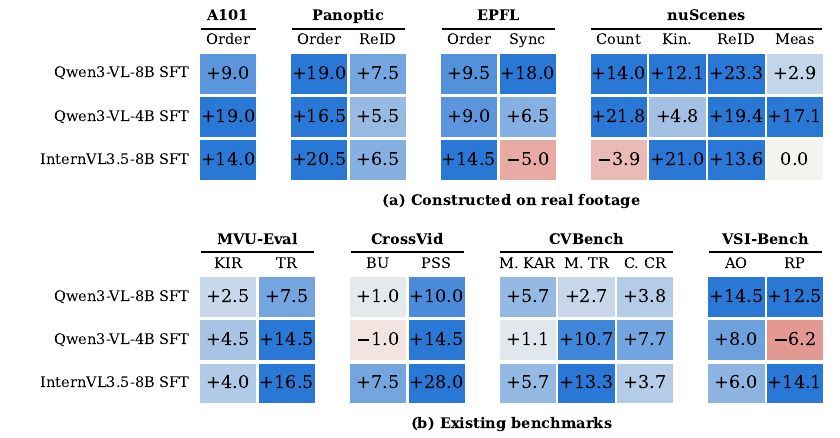}
    \caption{\newtext{\textbf{Real-video results after eight-task SYNCR SFT.} Each cell shows the accuracy change, in percentage points, from that model's own base on the same items. Blue indicates a gain and red a loss. Panel (a) uses SYNCR-format questions on real footage, while panel (b) shows individual tasks from existing benchmarks. Table~\ref{tab:fig5_absolute} gives the absolute scores and item counts behind every cell.}}
    \label{fig:sim2real_transfer}
\end{figure}

\newtext{\textbf{Ordering gains transfer to real footage.} We apply SYNCR's \taskOrder\ format to Assembly101~\citep{sener2022assembly101} and CMU Panoptic~\citep{joo2015panoptic}. Each fixed, answer-balanced set has 200 items formed by shuffling four gapped segments from a 20-second window; chronology supplies the label. Assembly101 uses static recordings and Panoptic individual camera streams, retaining within-stream ordering across new visual domains (App.~\ref{app:real_syncr}). Eight-task SFT improves Qwen3-VL-8B by 9.0 and 19.0 points on the two sets, InternVL3.5-8B by 14.0 and 20.5, and Qwen3-VL-4B by 19.0 and 16.5 (Fig.~\ref{fig:sim2real_transfer}). All six gains have paired McNemar $p\leq0.015$; the corrected Panoptic Qwen3-VL-8B result uses a conservative bound. Item-level intervals do not account for repeated windows from one recording (App.~\ref{app:real_syncr}).
}

\newtext{\textbf{Cross-camera counting transfer remains uncertain.} On a nuScenes~\citep{caesar2020nuscenes} set asking for distinct objects across six surround cameras, every checkpoint scores higher with six views than with one. The six-versus-one gap rises from 7.8 to 17.3 points for the first Qwen3-VL-8B SFT seed, but an increased gap is not established for its other seeds or InternVL3.5-8B. Additional views may expose more objects without improving cross-camera deduplication; absolute accuracy remains low (App.~\ref{app:real_syncr}).}

\textbf{Transfer on existing benchmarks.} On MVU-Eval~\citep{peng2025mvu} Temporal Reasoning and CrossVid~\citep{li2026crossvid} Procedural Step Sequencing, 8-task SFT gains 7.5 and 10.0 points for Qwen3-VL-8B, and 16.5 and 28.0 for InternVL3.5-8B, each relative to its own base. The Qwen gains persist when the same CrossVid items are answered in letter-string or prose formats absent from SYNCR training ($+19.0$ and $+8.0$ points; App.~\ref{app:realvalidity}). 
On VSI-Bench~\citep{yang2025thinking}, appearance order improves from 54.0\% to 68.5\% for Qwen3-VL-8B and from 46.0\% to 52.0\% for InternVL3.5-8B; route planning also rises in both families. These are independently defined single-video questions, so the appearance-order result tests transfer beyond both SYNCR's construction and the cross-video setting, rather than demonstrating cross-camera alignment.

\subsection{Controls and Robustness Checks}
\label{sec:training_controls}

\textbf{Incorrect-label supervision does not reproduce the gains.} A placebo adapter trained on the same examples, inputs, and options with random wrong labels scores below base on every SYNCR task and loses 7.5--35.0 points on five real-benchmark tasks. On MVU-Eval Temporal Reasoning, it scores 43.0\% versus 68.0\% for base and 75.5\% for correct SFT. This supports the role of correct labels; alternative answer formats are tested separately (Apps.~\ref{app:placebo} and~\ref{app:realvalidity}).

\textbf{The tested spatial baseline does not reproduce temporal transfer.} A matched-budget SAT~\citep{ray2024sat} arm, with frames rendered as clips, lowers MVU-Eval Temporal Reasoning and CrossVid Procedural Step Sequencing by 8.5 and 17.0 points. This result concerns that alternative under our format (App.~\ref{app:supervision_arms}).

\textbf{Training gains are larger with video input.} Removing videos leaves 8-task SFT near chance on seven tasks; \taskRoute\ is the exception at 39.5\%. The mean gain is 28.9 points with videos, and the full-video versus no-video gain contrast is 27.1 points. This contrast depends on video availability under the ablation but does not isolate a causal visual contribution (App.~\ref{app:video_attr}).

\textbf{Main gains persist across seeds and multiple-testing correction.} Three 8-task SFT seeds average 61.6\%, 61.8\%, and 64.6\% on SYNCR (task-level SD at most 5.1 points). A second seed gains 10.0 and 18.5 points on Assembly101 and Panoptic ordering, versus 9.0 and 19.0 for the first (App.~\ref{app:seeds}). Seven of eight SYNCR gains and both existing temporal-benchmark gains survive Benjamini--Hochberg correction (App.~\ref{app:multiplicity}).

\section{Related Work}
\label{sec:related}

\textbf{Multi-video evaluation.}
CVBench~\citep{zhu2025cvbench}, MVU-Eval~\citep{peng2025mvu}, CrossVid~\citep{li2026crossvid} and MVPBench~\citep{bai2026mvpbench} extend single-video evaluation~\citep{fu2025video,zhou2024mlvu,li2024mvbench} to questions that relate evidence across videos, and GameplayQA~\citep{wang2026gameplayqa} does so for synchronized agent viewpoints in virtual environments. Adjacent lines of work relate observations without a cross-video question format: static multi-view understanding~\citep{yeh2026allangles}, egocentric--exocentric activity~\citep{grauman2024egoexo4d} and video-based spatial memory~\citep{yang2025thinking}. Several benchmarks also check dependence on visual input (Table~\ref{tab:benchmark_comparison}). SYNCR's distinction is to couple simulator-state labels and shared generators with matched training data, evidence controls, and tests of transfer.

\textbf{Learning from simulated video data.}
Simulation already supplies supervision for single-video skills: Chain-of-Frames~\citep{ghazanfari2025chain} trains frame-grounded reasoning on CLEVRER, SAT~\citep{ray2024sat} and SIMS-V~\citep{brown2025simsv} train on spatial questions, and ST-VLM~\citep{ko2025stvlm} on kinematics, building on simulation-to-real transfer~\citep{tobin2017domain}. None of it supervises relating evidence across separate streams. \newtext{In our matched-example comparison, SAT frames rendered as clips did not reproduce SYNCR's temporal-transfer gains; this result concerns one alternative dataset and training recipe.} SYNCR trains on simulated data for cross-video reasoning, and couples that supervision with controls, establishing a framework in which a capability can be diagnosed, taught, and its transfer to real video measured under one set of task definitions.
\vspace{-3mm}
\section{Conclusion}

SYNCR connects simulator-grounded evaluation with targeted supervision across eight cross-video reasoning tasks. Across 22 models, performance is uneven, with persistent difficulties in physical comparison and scene integration. On matched 200-item sets per task, eight-task SFT raises Qwen3-VL-8B's mean accuracy from 32.6\% to 61.6\%, and several gains persist when task structure or video source changes. On real footage, the clearest replicated transfer is chronological ordering: three checkpoints gain 9.0--20.5 points on the constructed Assembly101 and Panoptic sets, with improvements also on existing temporal benchmarks. The results identify a skill that transfers from synthetic supervision while showing that transfer varies substantially by task.

\textbf{Limitations.}
SYNCR uses visual-only multiple-choice questions, mostly from simulation; simulator labels do not guarantee that every answer is visible in RGB. Real-footage ordering uses clips from one source stream, and item-level intervals ignore repeated recordings. Transfer on other real-footage tasks remains uncertain. RL findings concern the tested GRPO settings.

\subsection*{AI use statement}

We used generative AI tools for manuscript editing and feedback on clarity, presentation, and the scope of claims. The pillar icons in Fig.~\ref{fig:syncr-teaser} were generated using Google Gemini 3 Flash for illustration. We reviewed AI-assisted content for accuracy and take responsibility for the final text, claims, and artifacts.

\subsection*{Ethics statement}

SYNCR benchmark videos are simulator-rendered: Habitat traverses HM3D scans of real indoor spaces~\citep{ramakrishnan2021habitat}, whereas Kubric and CLEVRER depict generated object scenes. Transfer evaluations use existing real-footage datasets; we did not collect new footage for them. The human baseline study (App.~\ref{app:human_baseline}) involved four graduate-student volunteers who answered multiple-choice questions about benchmark videos; it did not collect personal, sensitive, demographic, or behavioral information beyond the selected answer choices. We do not anticipate direct harms from this benchmark; as with other video-understanding evaluations, downstream systems trained or evaluated on it should still be assessed for fairness and safety before deployment.

\subsection*{Reproducibility statement}

The SYNCR evaluation and training metadata, together with the Habitat and Kubric video assets, are available at \url{https://huggingface.co/datasets/CrossVideoReasoning/SYNCR}. CLEVRER videos must be obtained from the original CLEVRER release, as described in the dataset README. We describe the SYNCR data-generation pipeline in Sec.~\ref{sec:data}, with full implementation details for each simulation engine (Habitat, Kubric, and CLEVRER) provided in App.~\ref{app:dataset_details}. Our evaluation protocol, prompt templates, and statistical-significance methodology are described in Sec.~\ref{sec:experimental_setup} and App.~\ref{app:evaluation_details}, and the human baseline procedure is detailed in App.~\ref{app:human_baseline}. \newtext{We report 95\% Wilson confidence intervals for accuracy and exact McNemar tests for paired comparisons, and App.~\ref{app:training_details} lists all training hyperparameters.}

\subsection*{Acknowledgements}

This paper is supported in part by the Army Research Office under grant number W911NF-21-1-0155 and by the New York University Abu Dhabi (NYUAD) Center for Artificial Intelligence and Robotics, funded by Tamkeen under the NYUAD Research Institute Award CG010. Additional support was provided by the NYU IT High Performance Computing resources, services, and staff expertise.
\newpage
\bibliography{literature}
\bibliographystyle{iclr2027_conference}
\newpage
\appendix

\color{black}
\renewcommand{\newtext}[1]{{\color{black}#1}}
\renewenvironment{redpart}{\color{black}}{}

\section{The SYNCR Framework: Additional Details}
\label{app:framework}

\newtext{This appendix supports Sec.~\ref{sec:framework}: how each simulation engine supplies videos and annotations for programmatic question construction, and the controls establishing that the resulting items are answerable from the videos rather than from their text.}

\begin{table}[htbp]
\centering
\caption{\textbf{Comparison with multi-video benchmarks.} Columns refer to MVU-Eval~\citep{peng2025mvu}, CVBench~\citep{zhu2025cvbench}, CrossVid~\citep{li2026crossvid}, MVPBench~\citep{bai2026mvpbench}, and GameplayQA~\citep{wang2026gameplayqa}. Counts cover each full benchmark as defined by its authors; MVPBench and GameplayQA also include tasks that do not require multiple videos. In the training row, 0 means no separate training QA set is reported. A checkmark means the cited study reports the property; --- means it does not report it. For automatic QA construction, a checkmark covers at least one task; $\dagger$ marks CrossVid's eight automatically generated tasks out of ten. Controls differ in protocol.}
\label{tab:benchmark_comparison}
\footnotesize
\setlength{\tabcolsep}{2pt}
\renewcommand{\arraystretch}{1.15}
\newcommand{\benchmarktick}{\textcolor{green!50!black}{$\checkmark$}}
\newcommand{\benchmarktickpartial}{\textcolor{green!50!black}{$\checkmark^{\dagger}$}}
\begin{tabular}{@{}>{\raggedright\arraybackslash}p{0.35\linewidth}*{4}{>{\centering\arraybackslash}p{0.09\linewidth}}>{\centering\arraybackslash}p{0.10\linewidth}>{\centering\arraybackslash}p{0.09\linewidth}@{}}
\toprule
\textbf{Reported property} & \shortstack{\textbf{MVU-}\\\textbf{Eval}} & \shortstack{\textbf{CV}\\\textbf{Bench}} & \shortstack{\textbf{Cross}\\\textbf{Vid}} & \shortstack{\textbf{MVP}\\\textbf{Bench}} & \shortstack{\textbf{Gameplay}\\\textbf{QA}} & \shortstack{\textbf{SYNCR}\\\textbf{(ours)}} \\
\midrule
Evaluation QA pairs & 1,824 & 1,000 & 9,015 & 5,050 & 2,365 & 4,000 \\
Training QA pairs & 0 & 0 & 0 & 0 & 0 & 15,960 \\
Task categories/subtasks & 8 & 15 & 10 & 14 & 15 & 8 \\
\midrule
Automatic QA construction & \benchmarktick & \benchmarktick & \benchmarktickpartial & --- & \benchmarktick & \benchmarktick \\
Simulator-state answer labels & --- & --- & --- & --- & --- & \benchmarktick \\
\midrule
Direct no-video control & \benchmarktick & --- & --- & --- & \benchmarktick & \benchmarktick \\
Single-video comparison & \benchmarktick & \benchmarktick & --- & \benchmarktick & --- & \benchmarktick \\
Question/option-only audit & --- & --- & --- & --- & --- & \benchmarktick \\
\midrule
Shared-generator training/evaluation & --- & --- & --- & --- & --- & \benchmarktick \\
Post-training task/source shift tests & --- & --- & --- & --- & --- & \benchmarktick \\
Post-training real-video transfer & --- & --- & --- & --- & --- & \benchmarktick \\
\bottomrule
\end{tabular}
\end{table}

\subsection{Dataset Generation Details}
\label{app:dataset_details}

\newtext{This subsection documents how each engine supplies input videos and annotations for programmatic question construction. We describe rendering, visibility filtering, answer computation, and distractor generation, with the exact task prompts shown below.}

\subsubsection{Habitat}
\label{app:habitat}
\begin{figure}[htbp]
    \centering
    \includegraphics[width=\linewidth,trim={0cm 4.5cm 0cm 0cm}]{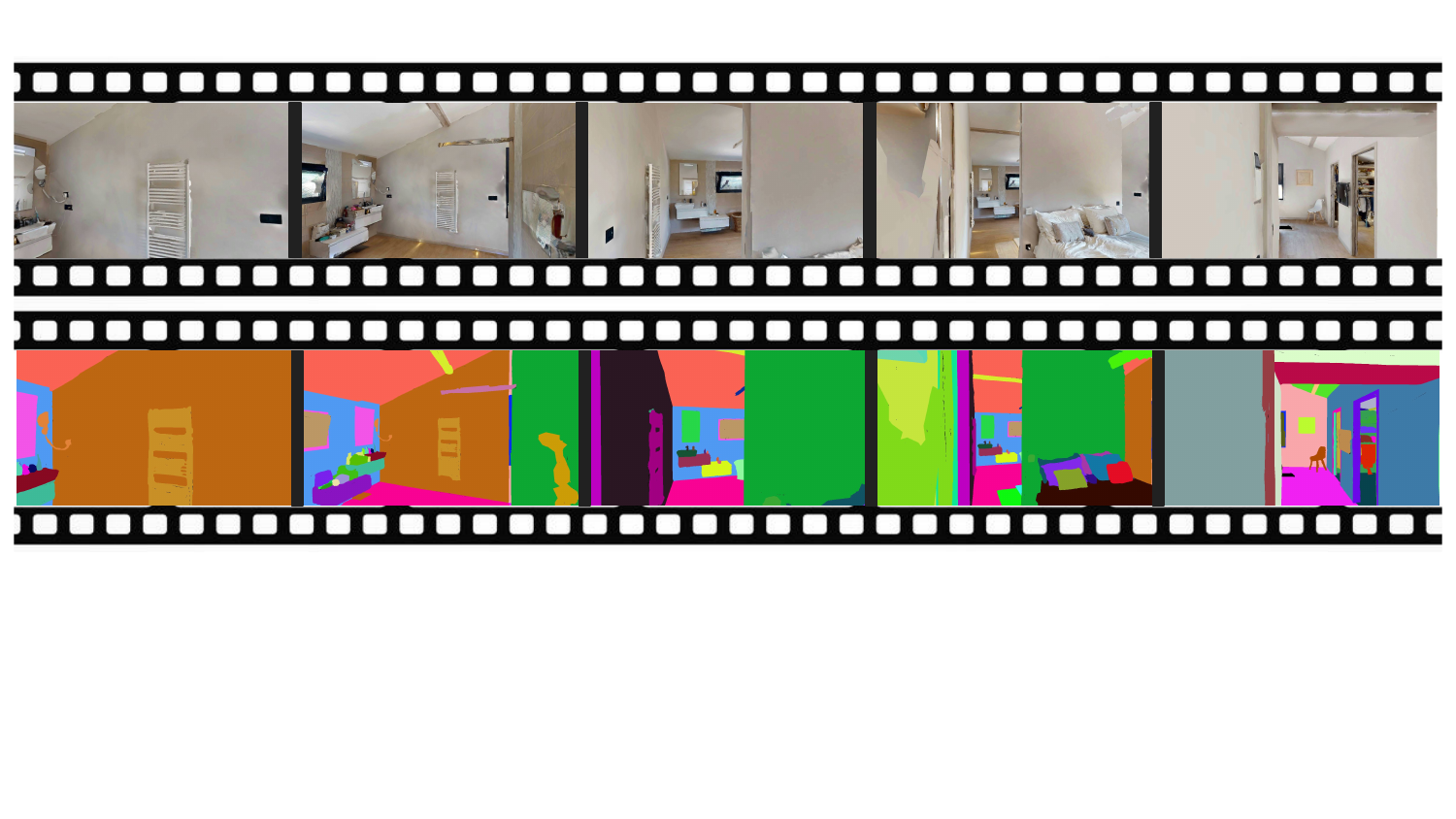}
   \caption{\textbf{Deterministic Ground Truth Generation in Habitat.} Habitat provides synchronized RGB, depth, and semantic observations for each rendered camera pose, from which SYNCR derives visibility, instance labels, and navigation structure.}
    \label{fig:habitat-semantic}
\end{figure}

\newtext{Habitat~\citep{savva2019habitat} with HM3D~\citep{ramakrishnan2021habitat} provides egocentric RGB observations, semantic instance annotations, and navigation meshes for large indoor environments. We use the 216 HM3D scenes with dense semantic annotations (Fig.~\ref{fig:habitat-semantic}).}

\textbf{Scene loading.} We load each HM3D scene's textured mesh, semantic annotations, and navigation mesh. Semantic annotations map instance IDs to normalized object names; the navigation mesh supports pathfinding and geodesic distances. We save RGB videos as MP4 files and semantic frame tensors separately. Only RGB is supplied to models; semantic tensors determine labels and filters. Scenes lacking semantic annotations or navigation meshes are skipped.

\newtext{\textbf{Trajectories.} Habitat's pathfinder returns a sparse shortest path for each sampled start--target pair. We remove duplicate waypoints, fit a cubic spline~\citep{virtanen2020scipy}, and snap interpolated points to the navigation mesh. Camera yaw follows a fixed look-ahead window to reduce jitter. Videos contain 100 frames at 5~FPS and end with a panoramic sweep. Geodesic sampling exposes spatially separated regions for \taskReID\ and \taskCount; topological routing samples connected paths and slices them into clips with continuous endpoints for \taskRoute.}

\newtext{\textbf{\taskcReID.} Each item contains a reference and a target video. Semantic tensors identify shared instances and their contiguous visibility windows. We retain objects visible for at least five consecutive frames, with sufficient total visibility across at least two views. The answer is the first contiguous window containing that instance in the target video. Distractors use answer windows from other items with similar reference windows and lie within the 20-second video, reducing cues from option position and timing (Fig.~\ref{fig:prompt_habitat_reid}).}

\begin{figure}[htbp]
\begin{spatialbox}[Habitat Object Re-identification Template]
You are provided with two videos recorded at {fps} FPS.
In Video 1, the '{object_name}' is clearly visible from {reference_time_window}.
Select the correct timestamp window for the FIRST appearance of the exact same '{object_name}' in Video 2.

Options:
A) {candidate_time_window_1}
B) {candidate_time_window_2}
C) {candidate_time_window_3}
D) {candidate_time_window_4}
\end{spatialbox}
\vspace{-2mm}
\caption{\textbf{Habitat Object Re-identification prompt template.} The model is given two Habitat videos and must identify the first timestamp window in Video 2 where the same semantic instance appears.}
\label{fig:prompt_habitat_reid}
\end{figure}

\newtext{\textbf{\taskcCount.} Each item contains three videos and asks for the number of distinct instances of two or three named categories. We compute totals from the union of semantic instance IDs across trajectories. Each counted instance must occupy at least 5\% of a frame in at least one video. We discard items with only one available video and categories with more than 10 instances. To reduce small-count priors, we retain totals of at least $k{+}3$ for $k$ categories and balance answers across totals. Options are four distinct values between $k$ and the answer plus four. We balance the answer's offset from the categories' typical total; audited text-only rules, including the smallest option and the value nearest that typical total, reach at most 32\% on 500 items (Fig.~\ref{fig:prompt_habitat_count}).}

\begin{figure}[htbp]
\begin{holisticbox}[Habitat Object Counting Template]
Based on the combined exploration in all {num_videos} videos,
how many distinct instances of '{object_name_1}' and '{object_name_2}'
[and '{object_name_3}'] are present in this scene in total?

Options:
A) {count_option_1}
B) {count_option_2}
C) {count_option_3}
D) {count_option_4}
\end{holisticbox}
\vspace{-2mm}
\caption{\textbf{Habitat Object Counting prompt template.} The model must aggregate observations across multiple Habitat videos and count unique semantic instances \newtext{of two or three named categories}.}
\label{fig:prompt_habitat_count}
\end{figure}

\newtext{\textbf{\taskcRoute.} Each item contains three clips of a route. We concatenate their route chunks, remove duplicated boundary nodes, and discard routes with fewer than three nodes. Questions use subpaths close to the full route length to encourage integration across clips; the answer is the shortest subpath between the designated regions. We describe each room by its three rarest objects, excluding background and structural terms. Distractors use rooms observed in the input videos and match the answer hop-count distribution, with answers split evenly between two and three or more hops. Distractors exclude direct start-to-goal paths and paths revisiting either endpoint, reducing obvious shortest-path cues (Fig.~\ref{fig:prompt_habitat_route}).}

\begin{figure}[htbp]
\begin{holisticbox}[Habitat Route Planning Template]
Based on the spatial layout shown across the videos, what is the shortest possible
path to travel from Region 1 <{start_region_description}> to Region 2 <{end_region_description}>?

Options:
A) {candidate_path_1}
B) {candidate_path_2}
C) {candidate_path_3}
D) {candidate_path_4}
\end{holisticbox}
\vspace{-2mm}
\caption{\textbf{Habitat Route Planning prompt template.} The model must infer the shortest navigable path between two regions by integrating partial spatial observations across multiple Habitat videos.}
\label{fig:prompt_habitat_route}
\end{figure}

\subsubsection{Kubric}
\label{app:kubric}

\newtext{\textbf{Simulation.} Kubric~\citep{greff2022kubric} renders dynamic object interactions with known camera parameters, object visibility, and 3D state. We simulate 10--15 objects in a $10 \times 10$ region with PyBullet~\citep{coumans2016pybullet} and render them with Blender's Cycles engine~\citep{blender2018} at $256 \times 256$ pixels, for 5~s at 12~FPS (60 frames).} The dataset is evenly split across two shape vocabularies: 1,000 scenes use the standard CLEVR shapes (cube, cylinder, sphere), and the remaining 1,000 use an expanded KuBasic suite of 11 primitives (including tori, gears, and teapots). To sustain motion, the floor is frictionless with a restitution of 0.5, and objects start with planar velocities sampled uniformly from $-4.0$ to $4.0$ units per second along the X and Y axes and zero vertical velocity.

\textbf{Camera rig.} A synchronized three-camera rig with a 35mm focal length gives distinct, partially overlapping views:
\begin{itemize}
\item \textbf{Camera 0 (Right):} Positioned at base coordinates (7.48, -6.50, 5.34) and angled to look at (4, 0, 0).
\item \textbf{Camera 1 (Left):} Positioned on the opposite side of the scene at (-7.48, 6.50, 5.34) and angled to look at (-4, 0, 0).
\item \textbf{Camera 2 (Front-Center):} Positioned lower and further back at (0.0, -13.5, 5.0), directly observing the origin (0, 0, 0).
\end{itemize}
We add a small uniform random offset to each camera's base position per scene to vary the extrinsics. Visibility matrices, 2D segmentation masks, and bounding boxes are recomputed per camera view, since they depend on the viewpoint. Object identities are aligned across views through shared asset identifiers, and object names are generated from their size, color, material, and shape.

\newtext{\textbf{\taskcSync.} We independently sample three 3-second crops from a scene's 5-second recording and randomly assign camera views to Videos~1--3. Videos~2 and~3 may start before or after Video~1; their relative crop-start times determine the answer. Distractor offsets use independent repetitions of the crop sampling, matching the answer distribution without systematic sign-flip or swap relations (Fig.~\ref{fig:prompt_kubric_sync}).}

\begin{figure}[htbp]
\begin{temporalbox}[Kubric Multi-Angle Synchronization Template]
You are provided with three videos (Video 1, Video 2, and Video 3) recorded at {fps} FPS.
Each video is exactly {crop_duration}.0 seconds long and shows the same physical event
captured simultaneously from three different camera angles.

Identify the exact temporal offset of the latter two videos. At what timestamp in Video 1's timeline
does the very first frame of Video 2 and Video 3 occur?
(Note: A negative timestamp means the video started before Video 1).

Options:
A) {offset_option_1}
B) {offset_option_2}
C) {offset_option_3}
D) {offset_option_4}
\end{temporalbox}
\vspace{-2mm}
\caption{\textbf{Kubric Multi-Angle Synchronization prompt template.} The model is given three cropped videos of the same physical event from different camera angles and must infer their temporal offsets.}
\label{fig:prompt_kubric_sync}
\end{figure}

\newtext{\textbf{\taskcMeasure.} Each item contains two camera views and is anchored at the first frame where a designated event object leaves one view. We discard scenes with duplicate object names. At the event frame, the target must be visible in both views and the answer in at least one. The answer minimizes simulator-derived 3D distance among visible objects. Distractors match the distributions of size, shape, material, and color relations to the target. All distractors exist in the scene, and no visible distractor is closer than the answer (Fig.~\ref{fig:prompt_kubric_spatial_measurement}).}

\begin{figure}[htbp]
\begin{spatialbox}[Kubric Spatial Measurement Template]
Observe the videos closely. At the exact moment the {event_object} completely exits the frame
in Video {video_number}, which object is physically closest to the {target_object} in the 3D space?

Options:
A) {candidate_object_1}
B) {candidate_object_2}
C) {candidate_object_3}
D) {candidate_object_4}
\end{spatialbox}
\vspace{-2mm}
\caption{\textbf{Kubric Spatial Measurement prompt template.} The model is given two camera views and must identify which candidate object is closest to a target object in simulator-derived 3D space.}
\label{fig:prompt_kubric_spatial_measurement}
\end{figure}

\subsubsection{CLEVRER}
\label{app:clevrer_generation}

CLEVRER~\citep{yi2019clevrer} provides simulated collision videos with frame-level annotations of object attributes, trajectories, velocities, visibility, and collision events. We reuse its original videos as model inputs and use the annotations only to construct and filter questions.

\newtext{\textbf{\taskcOrder.} We split each 128-frame, 25-FPS video into $k{=}4$ equal-length segments separated by random gaps of 0.3--0.6~s, starting at a random offset, so that no two segments share a boundary frame. The clips are shuffled, and the answer is their chronological order. Every ordering is equally frequent as an answer, and distractor orderings are sampled in proportion to answer frequencies (Fig.~\ref{fig:prompt_clevrer_order}).}

\begin{figure}[htbp]
\begin{temporalbox}[CLEVRER Sequential Ordering Template]
These {num_segments} videos are segments of a single continuous event,
but they are shown out of order. What is the correct chronological order of these video segments?

Options:
A) {candidate_order_1}
B) {candidate_order_2}
C) {candidate_order_3}
D) {candidate_order_4}
\end{temporalbox}
\vspace{-2mm}
\caption{\textbf{CLEVRER Sequential Ordering prompt template.} The model is given shuffled temporal segments from one continuous CLEVRER event and must recover their original chronological order.}
\label{fig:prompt_clevrer_order}
\end{figure}

\textbf{\taskcKinematic.} Each item has two videos. We compute each visible object's instantaneous speed as the norm of its 3D velocity and keep only pairs in which both the cross-video maximum velocities and the within-video top-two velocities differ by at least $\tau=0.25$. The options are the two fastest objects of each video, described by video index, color, material, and shape. \newtext{Answer letters are balanced in the evaluation set; in the training set, video pairs are non-consecutive and each video is used at most twice} (Fig.~\ref{fig:prompt_clevrer_kinematic}).

\begin{figure}[htbp]
\begin{comparativebox}[CLEVRER Kinematic Comparison Template]
Given two videos of CLEVRER physical interactions, identify which object has the highest peak velocity.

Options:
A) {candidate_video_object_1}
B) {candidate_video_object_2}
C) {candidate_video_object_3}
D) {candidate_video_object_4}
\end{comparativebox}
\vspace{-2mm}
\caption{\textbf{CLEVRER Kinematic Comparison prompt template.} The model must compare object motion across two CLEVRER videos and select the object with the simulator-derived highest peak velocity.}
\label{fig:prompt_clevrer_kinematic}
\end{figure}

\textbf{\taskcNumerical.} We sample $k$ distinct videos, count their annotated collisions, and ask for the video with the most collisions and its difference from the second-highest count. Ties use the answer ``most and second most have the same number of collisions -- 0'', and non-tie distractors combine wrong video identities, nearby wrong differences, and a tie option. \newtext{Each item has five options. Items are balanced across ties and differences of one, two, and three or more, with the winning video balanced; ties make up one fifth of the items, so always choosing the tie option scores chance (20\%), and no video is shared between evaluation items} (Fig.~\ref{fig:prompt_clevrer_numerical}).

\begin{figure}[htbp]
\begin{comparativebox}[CLEVRER Numerical Comparison Template]
Compare the total number of collisions across the {num_videos} videos.
Which video has the most collisions, and what is the numerical difference
in collisions between this video and the video with the second most collisions?

Select the correct option below.

Options:
A) {collision_option_1}
B) {collision_option_2}
C) {collision_option_3}
D) {collision_option_4}
E) {collision_option_5}
\end{comparativebox}
\vspace{-2mm}
\caption{\textbf{CLEVRER Numerical Comparison prompt template.} The model must compare collision counts across CLEVRER videos and identify both the video with the most collisions and the count difference.}
\label{fig:prompt_clevrer_numerical}
\end{figure}

\subsection{Dataset Statistics}
\label{app:dataset_stats}

\newtext{Table~\ref{tab:dataset_stats} gives the per-task composition of the benchmark and training set: the simulation engine, the number of distinct video files and question--answer pairs in each split, the number of videos per question, and the length of each clip shown to the model. Training videos are disjoint from evaluation videos.}

\begin{table}[!htbp]
    \centering
    \caption{\textbf{SYNCR dataset statistics.} Unique videos count distinct video files; Videos/Inp.\ is the number of videos per question; Clip Len.\ is the length of each clip shown to the model. Per-task counts are the distinct videos used by that task; the total is the union over tasks, which is smaller than their sum because some videos are shared between tasks. No video is shared between the evaluation and training splits. Total unique-video counts deduplicate files reused across tasks; the per-task counts sum to 5,688 for evaluation and 19,872 for training.}
    \label{tab:dataset_stats}
    \resizebox{\textwidth}{!}{%
    \begin{tabular}{ll|c|cc|cc|cc}
        \toprule
         & & & \multicolumn{2}{c|}{\textbf{Evaluation}} & \multicolumn{2}{c|}{\textbf{Training}} & & \\
        \textbf{Category} & \textbf{Task} & \textbf{Simulation Engine} & \textbf{Unique Videos} & \textbf{QA Pairs} & \textbf{Unique Videos} & \textbf{QA Pairs} & \textbf{Videos/Inp.} & \textbf{Clip Len. (s)} \\
        \midrule
        \rowcolor{temporalGreen} & Sequential Ordering & CLEVRER & 500 & 500 & 2,000 & 2,000 & 4 & 0.87 \\
        \rowcolor{temporalGreen} \multirow{-2}{2.5cm}{\ta} & Multi-Angle Synchronization & Kubric & 1,500 & 500 & 4,119 & 2,000 & 3 & 3.0 \\
        \midrule
        \rowcolor{entityYellow} & Object Re-identification & Habitat & 72 & 500 & 370 & 2,000 & 2 & 20.0 \\
        \rowcolor{entityYellow} \multirow{-2}{2.5cm}{\st} & Spatial Measurement & Kubric & 558 & 500 & 2,385 & 2,000 & 2 & 5.0 \\
        \midrule
        \rowcolor{comparativePeach} & Numerical Comparison & CLEVRER & 1,000 & 500 & 2,969 & 2,000 & 2 & 5.12 \\
        \rowcolor{comparativePeach} \multirow{-2}{2.5cm}{\cor} & Kinematic Comparison & CLEVRER & 837 & 500 & 2,677 & 2,000 & 2 & 5.12 \\
        \midrule
        \rowcolor{holisticBlue} & Object Counting & Habitat & 354 & 500 & 1,404 & 1,960 & 3 & 20.0 \\
        \rowcolor{holisticBlue} \multirow{-2}{2.5cm}{\hs} & Route Planning & Habitat & 867 & 500 & 3,948 & 2,000 & 3 & 20.0 \\
        \midrule
        \multicolumn{2}{l}{\textbf{Total / Average}} & \textbf{-} & \textbf{4,827} & \textbf{4,000} & \textbf{14,956} & \textbf{15,960} & \textbf{-} & \textbf{9.89} \\
        \bottomrule
    \end{tabular}
    }
\end{table}

\newtext{\subsection{Benchmark Validity}
\label{app:sanity}

This subsection assesses how task performance depends on text and on single-video evidence. Fig.~\ref{fig:validity_checks} summarizes the zero-shot controls; Table~\ref{tab:validity_controls} gives exact values. Answer letters and categories are balanced, and answers are not systematically the shortest or longest option. Evaluation videos are absent from training; the 33 Habitat evaluation scenes are disjoint from the 138 training scenes.

\subsubsection{Preventing Temporal Metadata Leakage}
\label{app:timestamp_leak}

Gemini receives clips cut to the requested crop window before upload, without source-video crop timestamps in the prompt. Qwen3-VL's per-frame timestamp markers are rebased so that each clip's first sampled frame is at $t{=}0$. Markers retain elapsed time within a clip but do not reveal its absolute position in the source recording. This prevents source timestamps from revealing the chronological order in \taskOrder\ or the crop offsets in \taskSync. Uncropped videos are unaffected. All reported evaluations use this protocol.}

\begin{table}[htbp]
  \centering
  \color{black}
  \captionsetup{font+=appendixblackfont,labelfont+=appendixblackfont}
  \caption{\newtext{\textbf{Video-dependence controls.} Accuracy (\%) with all videos, with one randomly selected video, and with no video; questions and options are unchanged.}}
  \label{tab:validity_controls}
  \adjustbox{max width=\linewidth}{
    \begin{tabular}{ll*{8}{c}}
    \toprule
    \textbf{Model} & & \textsc{Sync} & \textsc{Order} & \textsc{ReID} & \textsc{Meas} & \textsc{Num} & \textsc{Kin} & \textsc{Count} & \textsc{Route} \\
    \midrule
    Chance & & 25.0 & 25.0 & 25.0 & 25.0 & 20.0 & 25.0 & 25.0 & 25.0 \\
    \cmidrule(lr){1-10}
    Qwen3-VL-8B & all videos & 31.5 & 55.0 & 43.5 & 27.0 & 27.0 & 21.0 & 30.5 & 25.5 \\
    & one video & 28.5 & 24.0 & 31.0 & 28.0 & 17.5 & 19.5 & 26.5 & 23.0 \\
    & no video & 27.5 & 28.5 & 27.0 & 24.0 & 22.5 & 26.5 & 27.0 & 22.5 \\
    \cmidrule(lr){1-10}
    Qwen3-VL-32B & all videos & 31.5 & 66.0 & 56.0 & 28.0 & 31.5 & 19.0 & 29.5 & 32.0 \\
    & one video & 27.5 & 24.5 & 39.5 & 34.5 & 20.0 & 20.5 & 30.5 & 26.5 \\
    & no video & 24.6 & 24.4 & 26.0 & 24.6 & 21.4 & 24.6 & 23.4 & 22.6 \\
    \cmidrule(lr){1-10}
    GPT-6 Astra & all videos & 88.0 & 100.0 & 88.0 & 52.0 & 60.0 & 20.0 & 48.0 & 60.0 \\
    & one video & 12.0 & 24.0 & 28.0 & 32.0 & 4.0 & 8.0 & 28.0 & 28.0 \\
    & no video & 28.0 & 24.0 & 20.0 & 26.0 & 20.0 & 16.0 & 20.0 & 24.0 \\
    \bottomrule
    \end{tabular}}

\end{table}

\newtext{\subsubsection{No-Video and Single-Video Controls}
\label{app:blind}
\label{app:controls}

Distractors are sampled to avoid relations among the options that identify the answer without video (App.~\ref{app:dataset_details}). Answers are balanced over option letters and task-specific answer categories. On both the first 200 and all 500 evaluation items, no text-only rule we audited exceeds 33\% accuracy on any task (chance is 25\%, or 20\% for \taskNumerical).

\paragraph{No-video control.}
We remove the videos while keeping the question and options unchanged (Table~\ref{tab:validity_controls}). The no-video rows use the first 200 items for Qwen3-VL-8B, all 500 for Qwen3-VL-32B and the first 100 for GPT-6~Astra, which is asked to pick the most likely option; the 95\% interval around chance is about $\pm$6 points at 200 items, $\pm$4 at 500 and $\pm$8.5 at 100. No model exceeds that interval on any task, including on all 500 items for Qwen3-VL-8B, Qwen3-VL-32B and LLaVA-OneVision-72B. GPT-6~Astra falls marginally below it on \taskKinematic\ (16\%), which reflects an attribute preference the items penalise rather than a text cue they supply.

\paragraph{Single-video control.}
We retain one randomly selected video per item, again keeping the question and options unchanged. The control uses the first 200 items per task for Qwen models and 100 for GPT-6~Astra. \taskOrder\ drops from 55.0\% to 24.0\% for Qwen3-VL-8B and from 100\% to 24\% for GPT-6. One task retains above-chance accuracy: Qwen3-VL-32B reaches 39.5\% on \taskReID\ with one video versus 56.0\% with all videos.}

\section{Evaluating Cross-Video Reasoning: Additional Details}
\label{app:evaluation}

\newtext{This appendix supports Sec.~\ref{sec:experiments}: the zero-shot evaluation protocol and reporting conventions, the item sets behind each comparison, the human reference, and analyses extending the main results.}

\subsection{Experimental Details}
\label{app:evaluation_details}
\textbf{Evaluation protocol.} Each JSONL record stores the question, options, ground-truth answer, and ordered video list. Questions have four options, except \taskNumerical, which has five. We preserve stored option order and prepend each visual input with its video index (Fig.~\ref{fig:prompt_eval}). Records with \texttt{start\_sec} and \texttt{end\_sec} specify temporal crops; other records use the full video. Crop timestamps are handled as described in App.~\ref{app:timestamp_leak}.

We score the model's final answer using the shared evaluation parser. When an answer marker is present, its final stated option is compared with the ground truth. Otherwise, the parser accepts an explicit option commitment, a single option-lettered answer line, or a response matching one option's text. Unparsed responses are incorrect. This common scoring policy reduces penalties for answer-format deviations without treating an enumeration of options as a prediction.

\newtext{Three tasks (\taskOrder, \taskKinematic, \taskNumerical) re-use the original public CLEVRER videos, which may appear in some checkpoints' pretraining data. The questions and options are newly generated, so this is not answer leakage, but visual familiarity with the rendering style could affect the zero-shot profile on those three tasks; the Habitat and Kubric videos are rendered for SYNCR and are not public.}

\newtext{Table~\ref{tab:frame_sampling} lists the sampling rate and approximate frame count for each task. These rates apply to the standard SYNCR evaluation; model-specific frame limits and GRPO sampling are stated separately.}

\begin{table}[htbp]
    \centering
    \caption{\textbf{Frame sampling by task.} Frame sampling rate (FPS) and approximate number of frames extracted per video for each SYNCR task, for the standard SYNCR evaluation.}
    \label{tab:frame_sampling}
    \adjustbox{max width=0.8\columnwidth}{
    \begin{tabular}{L{46mm} C{30mm} C{16mm} C{22mm}}
        \toprule
        \textbf{Task} & \textbf{Pillar} & \textbf{FPS} & \textbf{Frames/video} \\
        \midrule
        \taskcSync  & \tacS & 12   & $\approx$36 \\
        \taskcOrder & \tacS & 25   & $\approx$22 \\
        \taskcReID  & \stcS & 2    & $\approx$40 \\
        \taskcMeasure & \stcS & 6  & $\approx$30 \\
        \taskcNumerical & \corcS & 6.25 & $\approx$32 \\
        \taskcKinematic & \corcS & 6.25 & $\approx$32 \\
        \taskcCount & \hscS & 2    & $\approx$40 \\
        \taskcRoute & \hscS & 2    & $\approx$40 \\
        \bottomrule
    \end{tabular}}
\end{table}

\begin{figure}[htbp]
\begin{promptbox}[Evaluation Prompt]
{question}

Options:
A) {option 1}
B) {option 2}
C) {option 3}
D) {option 4}
[E) {option 5}]

Format Instructions:
Provide your final answer in its own line and using exactly this format:
Answer: <option_alphabet>) <full_option_text>
\end{promptbox}
\vspace{-2mm}
\caption{\textbf{Prompt for evaluation.} We use the same multiple-choice evaluation prompt for all SYNCR tasks. \newtext{The fifth option appears only for \taskNumerical.}}
\label{fig:prompt_eval}
\end{figure}

\textbf{Statistical uncertainty.} \newtext{We report item-sampling uncertainty using 95\% Wilson score intervals~\citep{wilson1927probable} (Table~\ref{tab:full_results}). At chance, interval half-widths are approximately 3.8 points for 500 items and 6.0 points for 200. Paired model comparisons use the exact two-sided McNemar test~\citep{mcnemar1947note} on discordant items, including trained models against their own bases. These intervals do not measure training-seed or API-generation variability. Reuse of videos and scenes across questions can also limit the effective independence of items.}

\newtext{Table~\ref{tab:full_results} reports the same measurements as Table~\ref{tab:main_results}, with intervals given so that small differences between models can be read against sampling noise. Best and second-best per column are bold and underlined, and chance is 25\% for every task except \taskNumerical\ (20\%).}

\begin{table}[htbp]
  \centering
  \color{black}
  \captionsetup{font+=appendixblackfont,labelfont+=appendixblackfont}
  \caption{\textbf{Full-set results with confidence intervals.} \newtext{Accuracy (\%) on all 500 items per task for each standard open-weight checkpoint, with 95\% Wilson confidence intervals.}}
  \label{tab:full_results}
  \adjustbox{max width=\textwidth}{
    \begin{tabular}{L{26mm}*{8}{C{17mm}}C{7mm}}
    \toprule
    \multirow{2}{*}{\textbf{Model}} & \multicolumn{2}{c}{\textbf{\tacS}} & \multicolumn{2}{c}{\textbf{\stcS}} & \multicolumn{2}{c}{\textbf{\corcS}} & \multicolumn{2}{c}{\textbf{\hscS}} & \multirow{2}{*}{\textbf{Avg}} \\
    \cmidrule(lr){2-3} \cmidrule(lr){4-5} \cmidrule(lr){6-7} \cmidrule(lr){8-9}
    & \textsc{Sync} & \textsc{Order} & \textsc{ReID} & \textsc{Meas}  & \textsc{Num} & \textsc{Kin} & \textsc{Count} & \textsc{Route} & \\
    \midrule
    \multicolumn{10}{l}{\textbf{Model Size $\leq$ 4B}} \\
    \midrule
    LLaVA-OV 0.5B & 25.8 $\pm$ 3.8 & 24.4 $\pm$ 3.8 & 26.4 $\pm$ 3.9 & 23.0 $\pm$ 3.7 & 16.0 $\pm$ 3.2 & 24.0 $\pm$ 3.7 & 25.4 $\pm$ 3.8 & 25.4 $\pm$ 3.8 & 23.8 \\
    InternVL3.5 1B & 24.4 $\pm$ 3.8 & 25.8 $\pm$ 3.8 & 24.6 $\pm$ 3.8 & 24.4 $\pm$ 3.8 & 20.0 $\pm$ 3.5 & 27.6 $\pm$ 3.9 & 22.8 $\pm$ 3.7 & 27.4 $\pm$ 3.9 & 24.6 \\
    InternVL3.5 2B & 26.8 $\pm$ 3.9 & 25.2 $\pm$ 3.8 & 26.0 $\pm$ 3.8 & 31.2 $\pm$ 4.0 & 26.8 $\pm$ 3.9 & \underline{29.6 $\pm$ 4.0} & 28.4 $\pm$ 3.9 & 28.0 $\pm$ 3.9 & 27.8 \\
    Qwen3-VL 2B & 24.2 $\pm$ 3.7 & 28.4 $\pm$ 3.9 & 33.6 $\pm$ 4.1 & 29.4 $\pm$ 4.0 & 21.4 $\pm$ 3.6 & 20.2 $\pm$ 3.5 & 25.6 $\pm$ 3.8 & 23.2 $\pm$ 3.7 & 25.8 \\
    InternVL3.5 4B & 26.4 $\pm$ 3.9 & 30.6 $\pm$ 4.0 & 26.2 $\pm$ 3.8 & 24.8 $\pm$ 3.8 & 29.2 $\pm$ 4.0 & 24.2 $\pm$ 3.7 & 23.6 $\pm$ 3.7 & 25.6 $\pm$ 3.8 & 26.3 \\
    Qwen3-VL 4B & 26.2 $\pm$ 3.8 & 49.4 $\pm$ 4.4 & 46.4 $\pm$ 4.4 & \underline{38.0 $\pm$ 4.2} & 26.6 $\pm$ 3.9 & 20.0 $\pm$ 3.5 & 28.8 $\pm$ 4.0 & 28.0 $\pm$ 3.9 & 32.9 \\
    \midrule
    \multicolumn{10}{l}{\textbf{Model Size 4B $<$ size $\leq$ 8B}} \\
    \midrule
    LLaVA-OV 7B & 27.2 $\pm$ 3.9 & 21.6 $\pm$ 3.6 & 26.6 $\pm$ 3.9 & 27.6 $\pm$ 3.9 & 20.8 $\pm$ 3.6 & 24.2 $\pm$ 3.7 & 26.0 $\pm$ 3.8 & 23.6 $\pm$ 3.7 & 24.7 \\
    Qwen2.5-VL 7B & 25.0 $\pm$ 3.8 & 36.8 $\pm$ 4.2 & 25.8 $\pm$ 3.8 & 29.8 $\pm$ 4.0 & 22.4 $\pm$ 3.6 & 21.2 $\pm$ 3.6 & 27.2 $\pm$ 3.9 & 25.6 $\pm$ 3.8 & 26.7 \\
    InternVL3 8B & 26.0 $\pm$ 3.8 & 35.4 $\pm$ 4.2 & 25.0 $\pm$ 3.8 & 29.6 $\pm$ 4.0 & 26.6 $\pm$ 3.9 & 26.0 $\pm$ 3.8 & 27.2 $\pm$ 3.9 & 20.0 $\pm$ 3.5 & 27.0 \\
    InternVL3.5 8B & 26.4 $\pm$ 3.9 & 39.4 $\pm$ 4.3 & 26.0 $\pm$ 3.8 & 31.4 $\pm$ 4.1 & 31.6 $\pm$ 4.1 & \textbf{29.8 $\pm$ 4.0} & 26.4 $\pm$ 3.9 & 24.0 $\pm$ 3.7 & 29.4 \\
    Qwen3-VL 8B & 29.6 $\pm$ 4.0 & 56.4 $\pm$ 4.3 & \underline{47.6 $\pm$ 4.4} & 32.6 $\pm$ 4.1 & 29.6 $\pm$ 4.0 & 17.2 $\pm$ 3.3 & \underline{30.2 $\pm$ 4.0} & 29.4 $\pm$ 4.0 & 34.1 \\
    \midrule
    \multicolumn{10}{l}{\textbf{Model Size $>$ 8B}} \\
    \midrule
    InternVL3.5 14B & \textbf{31.0 $\pm$ 4.0} & 43.6 $\pm$ 4.3 & 31.2 $\pm$ 4.0 & 36.6 $\pm$ 4.2 & 32.2 $\pm$ 4.1 & 27.0 $\pm$ 3.9 & 25.8 $\pm$ 3.8 & 25.0 $\pm$ 3.8 & 31.6 \\
    Qwen2.5-VL 32B & 27.6 $\pm$ 3.9 & 30.2 $\pm$ 4.0 & 32.8 $\pm$ 4.1 & 30.2 $\pm$ 4.0 & 23.6 $\pm$ 3.7 & 21.8 $\pm$ 3.6 & 28.8 $\pm$ 4.0 & 31.2 $\pm$ 4.0 & 28.3 \\
    Qwen3-VL 32B & \underline{30.0 $\pm$ 4.0} & \underline{64.4 $\pm$ 4.2} & \textbf{56.2 $\pm$ 4.3} & 35.2 $\pm$ 4.2 & \underline{33.6 $\pm$ 4.1} & 19.4 $\pm$ 3.5 & 28.6 $\pm$ 3.9 & \underline{32.4 $\pm$ 4.1} & 37.5 \\
    InternVL3.5 38B & 26.0 $\pm$ 3.8 & 44.4 $\pm$ 4.3 & 34.6 $\pm$ 4.2 & \textbf{39.0 $\pm$ 4.3} & \textbf{41.0 $\pm$ 4.3} & 26.2 $\pm$ 3.8 & 30.0 $\pm$ 4.0 & \textbf{32.8 $\pm$ 4.1} & \underline{34.3} \\
    LLaVA-OV 72B & 21.8 $\pm$ 3.6 & 26.0 $\pm$ 3.8 & 31.2 $\pm$ 4.0 & 30.0 $\pm$ 4.0 & 22.6 $\pm$ 3.7 & 18.6 $\pm$ 3.4 & 27.4 $\pm$ 3.9 & 23.6 $\pm$ 3.7 & 25.2 \\
    Qwen2.5-VL 72B & 27.4 $\pm$ 3.9 & \textbf{66.8 $\pm$ 4.1} & 31.8 $\pm$ 4.1 & 31.6 $\pm$ 4.1 & 33.0 $\pm$ 4.1 & 24.4 $\pm$ 3.8 & \textbf{30.4 $\pm$ 4.0} & 25.8 $\pm$ 3.8 & 33.9 \\
    \bottomrule
    \end{tabular}%
  }
\end{table}

\begin{figure}[htbp]
    \centering
    \includegraphics[width=0.55\linewidth]{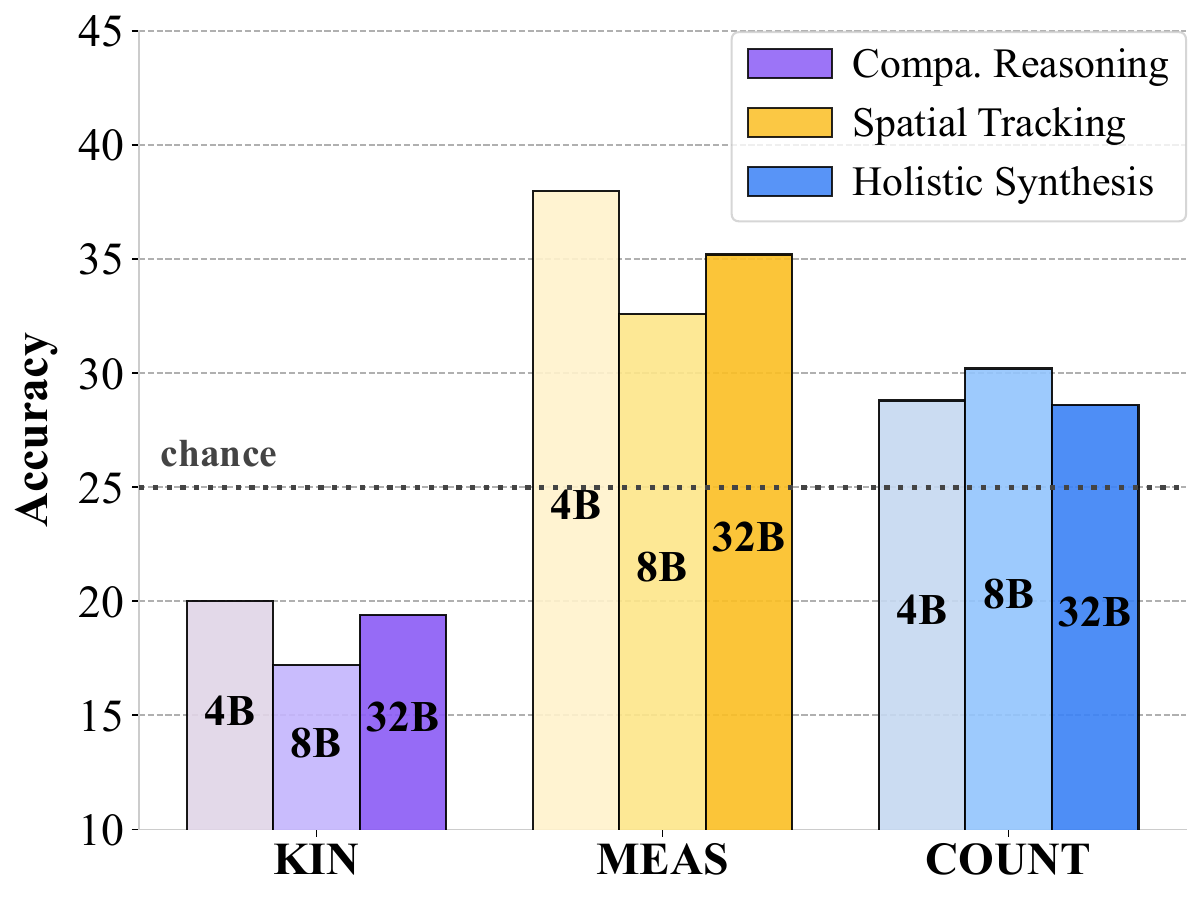}
    \caption{\newtext{\textbf{Tasks without consistent scaling gains} within Qwen3-VL (4B, 8B, 32B).}}
    \label{fig:non_scaling_tasks}
\end{figure}

\newtext{\paragraph{Task-level scaling.} Fig.~\ref{fig:non_scaling_tasks} isolates the three tasks that show no consistent improvement within Qwen3-VL at 4B, 8B and 32B: \taskKinematic, \taskMeasure\ and \taskCount.}

\paragraph{Compute and Hardware.} \newtext{Open-weight evaluations and all fine-tuning runs were executed on single compute nodes with NVIDIA A100 (80GB) or H100 (80GB) GPUs, using tensor parallelism for models that do not fit on one GPU.}

\subsection{Human Baseline Details}
\label{app:human_baseline}

Four graduate-student volunteers with machine-learning or computer-vision experience answered the first 100 items per task used for proprietary evaluation (800 items each), given only RGB videos, questions and options, with no simulator annotations, labels or generation metadata. We aggregate by majority vote, score two-way ties incorrect and weight tasks equally, giving the reported 89.5\% average.

\newtext{We report four-person majority-vote accuracy rather than individual human accuracy. Performance reaches 68.0\% on \taskKinematic\ and 76.0\% on \taskMeasure, indicating that these tasks remain challenging under the viewing protocol. Although simulator annotations provide precise ground-truth labels, these results do not establish that every answer is recoverable from the supplied RGB inputs.}

\begin{figure}[htbp]
    \centering
    \includegraphics[width=0.6\linewidth]{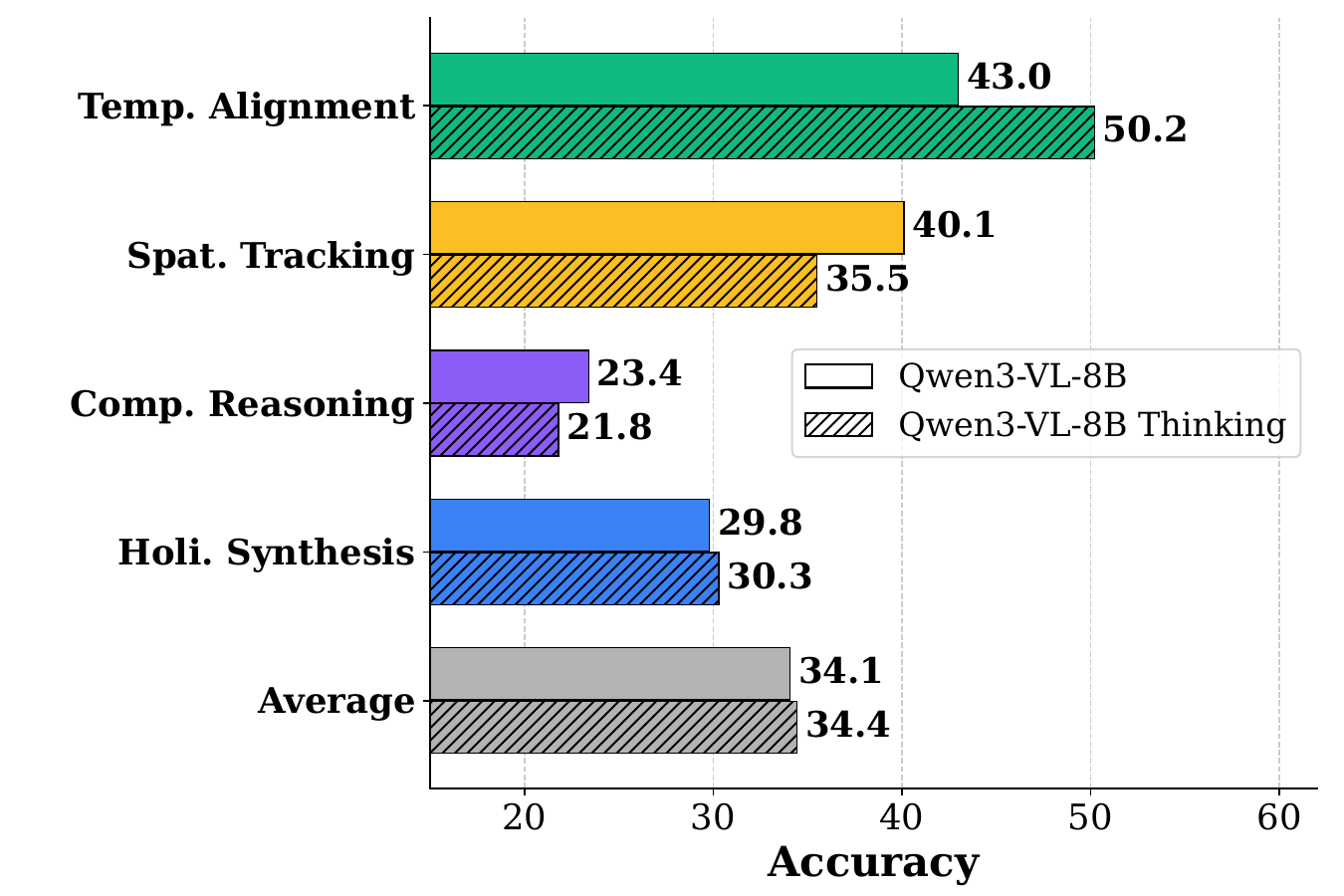}
    \caption{\textbf{Effect of reasoning-specialized post-training.}}
    \label{fig:qwen_thinking_checkpoint}
\end{figure}

\newtext{\subsection{Additional Zero-Shot Analyses}
\label{app:zeroshot}

\subsubsection{Reasoning-Specialized Checkpoints}
\label{app:thinking}
Reasoning-specialized post-training has uneven task-level effects. We compare Qwen3-VL-8B-Thinking with Instruct on the same 500 items per task (Fig.~\ref{fig:qwen_thinking_checkpoint}). On \taskOrder, Thinking reaches 76.6\% against Instruct's 56.4\%; on \taskSync, the scores are 23.8\% and 29.6\%, respectively. Temporal Alignment rises from 43.0\% to 50.2\% and the overall average from 34.1\% to 34.4\%, while Spatial Tracking falls from 40.1\% to 35.5\% and Comparative Reasoning from 23.4\% to 21.8\%. These descriptive differences do not isolate the effect of deliberation because the checkpoints differ in post-training.}

\newtext{\subsubsection{Agreement with a Real-Video Benchmark}
\label{app:construct_validity}

The controls in Sec.~\ref{sec:sanity} are internal to SYNCR: they show that items depend on the intended visual evidence, not that a checkpoint scoring better on SYNCR is better at real multi-video reasoning. We test that against a criterion SYNCR never touches, evaluating the same seventeen standard open-weight checkpoints on MVU-Eval Temporal Reasoning and Comparison through the identical harness (Table~\ref{tab:construct_validity}). SYNCR's ranking reproduces both benchmarks' rankings ($\rho = 0.855$ and $0.917$, both $p < 0.001$), and does so beyond what model size explains, since partialling on $\log_{10}$ parameters leaves $0.779$ and $0.870$. It is not carried by one family, as the seven Qwen checkpoints alone give $0.901$ and $0.927$, nor by one checkpoint, as leave-one-out leaves $\rho$ within $[0.817, 0.901]$ and $[0.905, 0.944]$; three checkpoints score below the 25\% chance level on MVU-Eval Comparison because they fail to emit a parseable option rather than because they reason worse, and excluding every below-chance row leaves $0.826$ and $0.860$. A general video-understanding factor would produce the same pattern, so the result shows that SYNCR does not rank models differently from real footage rather than that it isolates cross-video reasoning; and the relationship is not task-specific.}

\begin{table}[htbp]
  \centering
  \color{black}
  \captionsetup{font+=appendixblackfont,labelfont+=appendixblackfont}
  \caption{\newtext{\textbf{SYNCR against real multi-video benchmarks.} Accuracy (\%) for the seventeen standard open-weight checkpoints on SYNCR (eight-task average) and on MVU-Eval Temporal Reasoning and Comparison. All rows use the shared zero-shot harness; Tables~\ref{tab:sim2real_transfer} and~\ref{tab:fig5_absolute} instead evaluate InternVL3.5-8B through its native video path (at most 16 frames per video), where its base Temporal Reasoning accuracy is 60.0\% rather than 56.0\%.}}
  \label{tab:construct_validity}
  \adjustbox{max width=\linewidth}{
    \begin{tabular}{lrrrr}
    \toprule
    \textbf{Checkpoint} & \textbf{Params (B)} & \textbf{SYNCR avg} & \textbf{MVU-Eval TR} & \textbf{MVU-Eval Comp.} \\
    \midrule
    LLaVA-OV 0.5B & 0.5 & 23.8 & 14.0 & 9.6 \\
    InternVL3.5 1B & 1 & 24.6 & 30.0 & 1.5 \\
    InternVL3.5 2B & 2 & 27.8 & 39.0 & 51.8 \\
    Qwen3-VL 2B & 2 & 25.8 & 42.5 & 56.3 \\
    InternVL3.5 4B & 4 & 26.3 & 46.0 & 60.7 \\
    Qwen3-VL 4B & 4 & 32.9 & 58.5 & 65.9 \\
    Qwen2.5-VL 7B & 7 & 26.7 & 55.0 & 61.5 \\
    LLaVA-OV 7B & 7 & 24.7 & 36.0 & 48.1 \\
    InternVL3 8B & 8 & 27.0 & 56.0 & 61.5 \\
    InternVL3.5 8B & 8 & 29.4 & 56.0 & 64.4 \\
    Qwen3-VL 8B & 8 & 34.1 & 68.0 & 68.9 \\
    InternVL3.5 14B & 14 & 31.6 & 52.5 & 67.4 \\
    Qwen2.5-VL 32B & 32 & 28.3 & 64.0 & 65.9 \\
    Qwen3-VL 32B & 32 & 37.5 & 70.5 & 71.1 \\
    InternVL3.5 38B & 38 & 34.3 & 55.0 & 65.2 \\
    Qwen2.5-VL 72B & 72 & 33.9 & 70.5 & 71.1 \\
    LLaVA-OV 72B & 72 & 25.2 & 36.0 & 23.7 \\
    \midrule
    \multicolumn{3}{l}{\textit{Spearman $\rho$ with SYNCR average}} & 0.855 & 0.917 \\
    \multicolumn{3}{l}{\textit{\quad partialling out $\log_{10}$ params}} & 0.779 & 0.870 \\
    \bottomrule
    \end{tabular}}
\end{table}

\section{Learning from Synthetic Supervision: Additional Results}
\label{app:training_results}

\newtext{This appendix supports Sec.~\ref{sec:learning}: training configurations and complete results, the controls examining supervision content, label correctness, and answer-format dependence, and the generalization and sim-to-real analyses.}

\newtext{\subsection{Training Details}
\label{app:training_details}

\textbf{Data.} All runs use the same SYNCR training set (Table~\ref{tab:dataset_stats}): 2,000 examples per task, except 1,960 for \taskCount. Inputs come from the CLEVRER training split, Kubric scenes excluded from evaluation, and Habitat training scenes. SFT uses the task-specific evaluation sampling rates; GRPO uses the frame configuration below. Training mixes contain 7,960 (4-task), 8,000 (4-task-alt), 11,960 (6-task), and 15,960 (8-task) examples.

\textbf{SFT.} We fine-tune Qwen3-VL-8B-Instruct, and the additional bases of App.~\ref{app:other_models} (InternVL3.5-8B and Qwen3-VL-4B) under the same recipe, with LoRA~\citep{hu2022lora} (rank 16, $\alpha{=}32$, dropout 0.05) and a frozen vision encoder, for one epoch at a learning rate of $5{\times}10^{-5}$ with 3\% warmup and an effective batch size of \newtext{8 (one example per GPU and four accumulation steps on each of two GPUs)}. Targets contain only the answer line.

\textbf{RL.} We use GRPO~\citep{shao2024deepseekmath} with the same LoRA configuration, 8 rollouts per prompt, completions of up to 1,024 tokens, a learning rate of $10^{-5}$, and a KL coefficient of $\beta{=}0.02$. The reward is 1 when the parsed answer letter is correct and 0 otherwise. Prompts use 16 frames per video at the evaluation resolution. 4-task RL and 8-task RL train for 1,000 steps and the single-task runs for 500, with the \taskKinematic\ and \taskNumerical\ pilots run longer (Table~\ref{tab:rl_single_task}). SFT$\rightarrow$RL starts from the merged 8-task SFT models.}

\newtext{\subsection{Full SYNCR Training Results}
\label{app:training_full}
\label{app:single_task_rl}
\label{app:other_models}

Table~\ref{tab:training_syncr} reports all multi-task SFT and RL recipes. The additional 4-task-alt mix trains on \taskOrder, \taskMeasure, \taskKinematic, and \taskRoute, taking the complementary task from each pillar to the 4-task mix. The 6-task mix holds out both Kubric tasks (\taskSync, \taskMeasure).

\textbf{Cross-task effects.} Of the 56 off-diagonal cells in Fig.~\ref{fig:transfer_matrix}, 13 improve by more than 1.5 points, 27 change by at most 1.5 points, and 16 decline by more than 1.5 points. This descriptive threshold is not a statistical significance criterion. All seven adapters trained without \taskSync\ lower it by 3.0--8.5 points; the seven adapters trained without \taskOrder\ change it by $-11.5$ to $+5.5$ points. \taskKinematic\ and \taskMeasure\ each improve after four other single-task adapters, by at most 9.0 and 4.5 points. All eight adapters raise average accuracy, by 0.1--5.5 points.

\textbf{Tasks excluded from training.}\label{sec:heldout_transfer} The 6-task mix excludes both Kubric tasks and reaches 28.0\% on \taskSync\ and 23.5\% on \taskMeasure\ (base 31.5\% and 27.0\%). Single-task adapters trained without \taskMeasure\ change it by 0.0--4.5 points. The 4-task mix reaches 29.5\% on held-out \taskMeasure\ and 24.5\% on held-out \taskKinematic\ (base 27.0\% and 21.0\%), while held-out \taskOrder\ falls from 55.0\% to 36.0\%. These comparisons provide limited evidence of transfer to tasks absent from a model's training mix.

\textbf{SFT--RL comparison.} RL from the base model reaches 43.0--47.0\% on \taskReID\ (base 43.5\%) and 23.0--29.0\% on \taskKinematic\ (base 21.0\%), against 71.5--75.5\% and 73.5--76.0\% for SFT. These findings are specific to GRPO with binary correctness rewards and the training configuration in App.~\ref{app:training_details}.}

\newtext{\textbf{Single-task RL.} Trained on one task for 500 steps, GRPO raises accuracy on six of the seven evaluated trained tasks --- most strongly \taskKinematic, 21.0\% to 48.5\% --- while \taskSync\ falls by 1.5 points (Table~\ref{tab:rl_single_task}; trained task bold, last column its SFT adapter, \taskNumerical\ evaluation incomplete). All stay below their SFT adapters, cross-task changes span $+9.0$ to $-9.0$ points, and multi-task RL never matches that gain (23.0--29.0\%).}

\newtext{\textbf{A second SFT seed.} We retrained the 8-task recipe with a different seed, which changes the LoRA initialization, the data order and the validation split. On the first 200 items per task it reaches 69.5, 98.0, 74.5, 46.0, 48.0, 72.0, 34.5 and 52.0\% (\textsc{Sync} to \textsc{Route}; average 61.8\%, against 61.6\% for the first seed). In a separate eleven-task evaluation on existing real-video benchmarks, it gains over the base model on ten, including $+10.0$ on MVU-Eval Temporal Reasoning, $+10.0$ on CrossVid PSS and $+10.6$ on CVBench Temporal Reasoning, and loses 2.0 points on CVBench Scene Recognition.}

\newtext{The 4-task mix contains \taskSync, \taskReID, \taskNumerical\ and \taskCount; 4-task-alt contains their complements; 6-task excludes the Kubric tasks (\taskSync, \taskMeasure); 8-task contains all tasks. RL rows use their named mixes, and SFT$\rightarrow$RL starts from 8-task SFT.}

\providecommand{\tn}{\rlap{$^\dagger$}}
\providecommand{\td}{\rlap{$^{\dagger\ddagger}$}}
\begin{table}[htbp]
  \centering
  \caption{\textbf{SYNCR accuracy after fine-tuning Qwen3-VL-8B.} Accuracy (\%) on matched 200-item sets per task, scored by the common answer parser. $^\dagger$ marks trained tasks; unmarked columns evaluate excluded tasks.}
  \label{tab:training_syncr}
  \adjustbox{max width=0.9\textwidth}{
    \begin{tabular}{L{40mm}*{8}{C{11mm}}C{9mm}}
    \toprule
    \multirow{2}{*}{\textbf{Model}} & \multicolumn{2}{c}{\textbf{\tacS}} & \multicolumn{2}{c}{\textbf{\stcS}} & \multicolumn{2}{c}{\textbf{\corcS}} & \multicolumn{2}{c}{\textbf{\hscS}} & \multirow{2}{*}{\textbf{Avg}} \\
    \cmidrule(lr){2-3} \cmidrule(lr){4-5} \cmidrule(lr){6-7} \cmidrule(lr){8-9}
    & \textsc{Sync} & \textsc{Order} & \textsc{ReID} & \textsc{Meas} & \textsc{Num} & \textsc{Kin} & \textsc{Count} & \textsc{Route} & \\
    \midrule
    Base (Qwen3-VL 8B) & 31.5 & 55.0 & 43.5 & 27.0 & 27.0 & 21.0 & 30.5 & 25.5 & 32.6 \\
    \midrule
    \multicolumn{10}{l}{\textbf{SFT, multi-task}} \\
    \midrule
    4-task \scriptsize(Sync, ReID, Num, Count) & \textbf{69.5}\tn & 36.0 & \textbf{74.5}\tn & 29.5 & \textbf{49.0}\tn & 24.5 & \textbf{32.0}\tn & 22.5 & 42.2 \\
    4-task-alt \scriptsize(Order, Meas, Kin, Route) & 24.0 & \textbf{98.5}\tn & 43.0 & \textbf{46.0}\tn & 31.5 & \textbf{76.0}\tn & 26.5 & \textbf{48.5}\tn & 49.3 \\
    6-task \scriptsize(no Kubric)            & 28.0 & 96.0\tn & 71.5\tn & 23.5 & 48.0\tn & 74.0\tn & 33.0\tn & 49.0\tn & 52.9 \\
    8-task                                   & 70.0\tn & 98.0\tn & 73.0\tn & 39.5\tn & 50.0\tn & 75.0\tn & 34.5\tn & 52.5\tn & 61.6 \\
    \midrule
    \multicolumn{10}{l}{\textbf{RL (GRPO) from the base model}} \\
    \midrule
    4-task \scriptsize(Sync, ReID, Num, Count) & 24.5\tn & 63.5 & 47.0\tn & 27.5 & 32.0\tn & 23.0 & 29.5\tn & 24.0 & 33.9 \\
    8-task \scriptsize(all eight)               & 29.0\tn & 79.0\tn & 43.0\tn & 26.5\tn & 23.5\tn & 28.0\tn & 28.5\tn & 23.5\tn & 35.1 \\
    8-task, seed 2                              & 26.0\tn & 75.5\tn & 47.0\tn & 29.0\tn & 34.0\tn & 29.0\tn & 28.5\tn & 25.5\tn & 36.8 \\
    \midrule
    \multicolumn{10}{l}{\textbf{SFT followed by RL (GRPO)}} \\
    \midrule
    8-task SFT $\rightarrow$ RL       & \textbf{78.0}\tn & 98.0\tn & \textbf{75.5}\tn & \textbf{40.0}\tn & 48.0\tn & 75.0\tn & \textbf{32.5}\tn & \textbf{53.0}\tn & \textbf{62.5} \\
    \bottomrule
    \end{tabular}%
  }
\end{table}

\newtext{In Fig.~\ref{fig:transfer_matrix} the trained task is bold, blue and orange mark gains and losses beyond 1.5 points, and $\Delta$ is the change in the eight-task average; the multi-task models are in Table~\ref{tab:training_syncr}.}

\begin{figure}[!htb]
  \centering
  \color{black}
  \captionsetup{font+=appendixblackfont,labelfont+=appendixblackfont}
  \renewcommand{\arraystretch}{1.0}
  \resizebox{\linewidth}{!}{
    \begin{tabular}{l|cc|cc|cc|cc|cc}
    \toprule
    & \multicolumn{2}{c|}{\textbf{\tacS}} & \multicolumn{2}{c|}{\textbf{\stcS}} & \multicolumn{2}{c|}{\textbf{\corcS}} & \multicolumn{2}{c|}{\textbf{\hscS}} & & \\
    \textbf{Trained on} $\downarrow$ \;/\; \textbf{Evaluated on} $\rightarrow$ & \textsc{Sync} & \textsc{Order} & \textsc{ReID} & \textsc{Meas} & \textsc{Num} & \textsc{Kin} & \textsc{Count} & \textsc{Route} & \textbf{Avg} & $\boldsymbol{\Delta}$ \\
    \midrule
    Base (Qwen3-VL 8B) & 31.5 & 55.0 & 43.5 & 27.0 & 27.0 & 21.0 & 30.5 & 25.5 & 32.6 & --- \\
    \midrule
    \textsc{Sync} only & \cellcolor{blue!15}\makecell{\textbf{37.0}\\[-2pt]{\scriptsize(+5.5)}} & \cellcolor{orange!14}\makecell{49.0\\[-2pt]{\scriptsize($-$6.0)}} & \makecell{45.0\\[-2pt]{\scriptsize(+1.5)}} & \makecell{28.5\\[-2pt]{\scriptsize(+1.5)}} & \makecell{26.5\\[-2pt]{\scriptsize($-$0.5)}} & \makecell{21.0\\[-2pt]{\scriptsize(+0.0)}} & \makecell{30.0\\[-2pt]{\scriptsize($-$0.5)}} & \makecell{24.5\\[-2pt]{\scriptsize($-$1.0)}} & 32.7 & +0.1 \\
    \textsc{Order} only & \cellcolor{orange!15}\makecell{26.0\\[-2pt]{\scriptsize($-$5.5)}} & \cellcolor{blue!40}\makecell{\textbf{98.5}\\[-2pt]{\scriptsize(+43.5)}} & \makecell{45.0\\[-2pt]{\scriptsize(+1.5)}} & \makecell{27.0\\[-2pt]{\scriptsize(+0.0)}} & \makecell{26.5\\[-2pt]{\scriptsize($-$0.5)}} & \cellcolor{blue!15}\makecell{26.5\\[-2pt]{\scriptsize(+5.5)}} & \cellcolor{orange!12}\makecell{28.5\\[-2pt]{\scriptsize($-$2.0)}} & \makecell{24.5\\[-2pt]{\scriptsize($-$1.0)}} & 37.8 & +5.2 \\
    \cmidrule(lr){1-11}
    \textsc{ReID} only & \cellcolor{orange!17}\makecell{23.0\\[-2pt]{\scriptsize($-$8.5)}} & \cellcolor{blue!14}\makecell{60.5\\[-2pt]{\scriptsize(+5.5)}} & \cellcolor{blue!32}\makecell{\textbf{73.0}\\[-2pt]{\scriptsize(+29.5)}} & \cellcolor{blue!13}\makecell{30.0\\[-2pt]{\scriptsize(+3.0)}} & \makecell{26.5\\[-2pt]{\scriptsize($-$0.5)}} & \cellcolor{blue!14}\makecell{25.5\\[-2pt]{\scriptsize(+4.5)}} & \cellcolor{blue!13}\makecell{33.0\\[-2pt]{\scriptsize(+2.5)}} & \cellcolor{orange!13}\makecell{22.5\\[-2pt]{\scriptsize($-$3.0)}} & 36.8 & +4.1 \\
    \textsc{Meas} only & \cellcolor{orange!13}\makecell{28.5\\[-2pt]{\scriptsize($-$3.0)}} & \cellcolor{blue!12}\makecell{58.0\\[-2pt]{\scriptsize(+3.0)}} & \makecell{45.0\\[-2pt]{\scriptsize(+1.5)}} & \cellcolor{blue!20}\makecell{\textbf{40.0}\\[-2pt]{\scriptsize(+13.0)}} & \cellcolor{orange!15}\makecell{21.5\\[-2pt]{\scriptsize($-$5.5)}} & \cellcolor{blue!13}\makecell{23.5\\[-2pt]{\scriptsize(+2.5)}} & \cellcolor{orange!13}\makecell{28.0\\[-2pt]{\scriptsize($-$2.5)}} & \cellcolor{orange!15}\makecell{20.5\\[-2pt]{\scriptsize($-$5.0)}} & 33.1 & +0.5 \\
    \cmidrule(lr){1-11}
    \textsc{Num} only & \cellcolor{orange!16}\makecell{24.0\\[-2pt]{\scriptsize($-$7.5)}} & \cellcolor{orange!19}\makecell{43.5\\[-2pt]{\scriptsize($-$11.5)}} & \makecell{45.0\\[-2pt]{\scriptsize(+1.5)}} & \cellcolor{blue!13}\makecell{30.0\\[-2pt]{\scriptsize(+3.0)}} & \cellcolor{blue!27}\makecell{\textbf{49.5}\\[-2pt]{\scriptsize(+22.5)}} & \cellcolor{blue!17}\makecell{30.0\\[-2pt]{\scriptsize(+9.0)}} & \makecell{31.0\\[-2pt]{\scriptsize(+0.5)}} & \makecell{24.0\\[-2pt]{\scriptsize($-$1.5)}} & 34.6 & +2.0 \\
    \textsc{Kin} only & \cellcolor{orange!16}\makecell{24.0\\[-2pt]{\scriptsize($-$7.5)}} & \makecell{55.5\\[-2pt]{\scriptsize(+0.5)}} & \makecell{44.0\\[-2pt]{\scriptsize(+0.5)}} & \cellcolor{blue!13}\makecell{30.0\\[-2pt]{\scriptsize(+3.0)}} & \makecell{26.5\\[-2pt]{\scriptsize($-$0.5)}} & \cellcolor{blue!40}\makecell{\textbf{73.5}\\[-2pt]{\scriptsize(+52.5)}} & \makecell{29.0\\[-2pt]{\scriptsize($-$1.5)}} & \cellcolor{orange!13}\makecell{22.5\\[-2pt]{\scriptsize($-$3.0)}} & 38.1 & +5.5 \\
    \cmidrule(lr){1-11}
    \textsc{Count} only & \cellcolor{orange!14}\makecell{27.0\\[-2pt]{\scriptsize($-$4.5)}} & \makecell{56.5\\[-2pt]{\scriptsize(+1.5)}} & \cellcolor{blue!12}\makecell{45.5\\[-2pt]{\scriptsize(+2.0)}} & \cellcolor{blue!14}\makecell{31.5\\[-2pt]{\scriptsize(+4.5)}} & \makecell{28.0\\[-2pt]{\scriptsize(+1.0)}} & \makecell{20.5\\[-2pt]{\scriptsize($-$0.5)}} & \cellcolor{blue!14}\makecell{\textbf{35.0}\\[-2pt]{\scriptsize(+4.5)}} & \makecell{26.0\\[-2pt]{\scriptsize(+0.5)}} & 33.8 & +1.1 \\
    \textsc{Route} only & \cellcolor{orange!17}\makecell{23.5\\[-2pt]{\scriptsize($-$8.0)}} & \cellcolor{blue!13}\makecell{59.5\\[-2pt]{\scriptsize(+4.5)}} & \makecell{43.5\\[-2pt]{\scriptsize(+0.0)}} & \makecell{27.5\\[-2pt]{\scriptsize(+0.5)}} & \cellcolor{orange!15}\makecell{21.5\\[-2pt]{\scriptsize($-$5.5)}} & \makecell{21.0\\[-2pt]{\scriptsize(+0.0)}} & \makecell{30.0\\[-2pt]{\scriptsize($-$0.5)}} & \cellcolor{blue!32}\makecell{\textbf{55.0}\\[-2pt]{\scriptsize(+29.5)}} & 35.2 & +2.6 \\
    \bottomrule
    \end{tabular}%
  }
  \caption{\textbf{Transfer matrix for single-task SFT.} \newtext{Accuracy (\%) of Qwen3-VL-8B after training on the row task, evaluated on the column task (200 items each), with the change from the base in parentheses.}}
  \label{fig:transfer_matrix}
\end{figure}

\begin{table}[htbp]
  \centering
  \color{black}
  \captionsetup{font+=appendixblackfont,labelfont+=appendixblackfont}
  \caption{\newtext{\textbf{Single-task RL.} Accuracy (\%) of Qwen3-VL-8B after 500 GRPO steps on one task, evaluated on all eight tasks (200 items each); the trained task is bold.}}
  \label{tab:rl_single_task}
  \adjustbox{max width=\linewidth}{
    \begin{tabular}{l*{9}{c}c}
    \toprule
    \textbf{Trained on} & \textsc{Sync} & \textsc{Order} & \textsc{ReID} & \textsc{Meas} & \textsc{Num} & \textsc{Kin} & \textsc{Count} & \textsc{Route} & \textbf{Avg} & \textbf{SFT} \\
    \midrule
    Base & 31.5 & 55.0 & 43.5 & 27.0 & 27.0 & 21.0 & 30.5 & 25.5 & 32.6 & --- \\
    \textsc{Sync} & \textbf{30.0} & 62.5 & 45.0 & 27.0 & 18.0 & 23.0 & 29.0 & 26.0 & 32.6 & 37.0 \\
    \textsc{Order} & 30.5 & \textbf{73.0} & 43.5 & 26.5 & 27.5 & 23.5 & 29.5 & 24.5 & 34.8 & 98.5 \\
    \textsc{ReID} & 25.5 & 60.0 & \textbf{44.5} & 27.5 & 22.5 & 23.0 & 29.5 & 24.5 & 32.1 & 73.0 \\
    \textsc{Meas} & 27.5 & 60.5 & 42.5 & \textbf{33.0} & 18.0 & 21.5 & 30.5 & 24.5 & 32.3 & 40.0 \\
    \textsc{Num} & --- & --- & --- & --- & --- & --- & --- & --- & --- & 49.5 \\
    \textsc{Kin} & 28.5 & 64.0 & 45.0 & 28.0 & 28.0 & \textbf{48.5} & 29.0 & 26.5 & 37.2 & 73.5 \\
    \textsc{Count} & 27.0 & 59.0 & 44.0 & 27.0 & 26.5 & 21.5 & \textbf{32.0} & 25.5 & 32.8 & 35.0 \\
    \textsc{Route} & 26.5 & 59.5 & 43.5 & 26.5 & 25.0 & 22.5 & 29.0 & \textbf{36.0} & 33.6 & 55.0 \\
    \bottomrule
    \end{tabular}}

\end{table}

\newtext{\textbf{Other base models.} Table~\ref{tab:cross_model_training} reports the 8-task SFT recipe on two additional bases. InternVL3.5-8B improves from 29.3\% to 62.8\% on matched 200-item sets, with gains on every task. Qwen3-VL-4B reaches 62.6\% on the first 200 items per task, comparable to the 61.6\% reached by the trained 8B model on those items. Its available zero-shot result is 32.9\% on all 500 items per task (Table~\ref{tab:main_results}); without its first-200 scores, we do not estimate a matched 4B training gain.}

\newtext{InternVL3.5-8B uses its native Hugging Face video path, which samples at most 16 frames per video, so it is compared only with its own base.}

\begin{table}[htbp]
  \centering
  \color{black}
  \captionsetup{font+=appendixblackfont,labelfont+=appendixblackfont}
  \caption{\newtext{\textbf{SFT on other base models.} SYNCR accuracy (\%) on the first 200 items per task; the trained tasks are bold. The Qwen3-VL-4B base was evaluated on all 500 items per task in Table~\ref{tab:main_results} and is omitted here because matched first-200 scores are unavailable.}}
  \label{tab:cross_model_training}
  \adjustbox{max width=\linewidth}{
    \begin{tabular}{l*{8}{c}c}
    \toprule
    \textbf{Model} & \textsc{Sync} & \textsc{Order} & \textsc{ReID} & \textsc{Meas} & \textsc{Num} & \textsc{Kin} & \textsc{Count} & \textsc{Route} & \textbf{Avg} \\
    \midrule
    \multicolumn{10}{l}{\textbf{InternVL3.5-8B}} \\
    Base & 23.5 & 42.5 & 33.0 & 25.5 & 27.5 & 33.0 & 29.0 & 20.5 & 29.3 \\
    8-task SFT & \textbf{85.0} & \textbf{98.5} & \textbf{62.0} & \textbf{40.0} & \textbf{51.0} & \textbf{71.5} & \textbf{35.0} & \textbf{59.0} & 62.8 \\
    \midrule
    \multicolumn{10}{l}{\textbf{Qwen3-VL-4B}} \\
    8-task SFT & \textbf{70.5} & \textbf{98.5} & \textbf{68.5} & \textbf{46.0} & \textbf{47.5} & \textbf{74.5} & \textbf{32.0} & \textbf{63.5} & 62.6 \\
    \bottomrule
    \end{tabular}}
\end{table}

\newtext{\subsection{Evaluation Without Answer Options}
\label{app:count_openended}

\paragraph{Counting without answer options.} \taskCount\ shows smaller multiple-choice gains than most other tasks. Exact-match scoring over four numeric options reports only whether a model lands on the right integer, so we also evaluate it open-ended, replacing the options with a request for a number and parsing the number from the response (Table~\ref{tab:count_openended}). MRA is Mean Relative Accuracy~\citep{yang2025thinking}; $\rho$ is the Spearman correlation between predicted and true counts, which unlike MRA is unaffected by a uniform scale bias. Scored on 200 items for the Qwen models and 500 for InternVL3.5-8B, eight-task SFT raises MRA in all three families and raises the rank correlation between predicted and true counts by 0.11 to 0.24, with paired bootstrap intervals excluding zero.}

\begin{table}[htbp]
  \centering
  \color{black}
  \captionsetup{font+=appendixblackfont,labelfont+=appendixblackfont}
  \caption{\newtext{\textbf{\taskCount\ without answer options.} Models answer with a number instead of choosing among four, scored by parsing the number from the response. Base and fine-tuned are scored on identical items.}}
  \label{tab:count_openended}
  \adjustbox{max width=\linewidth}{
    \begin{tabular}{lccc}
    \toprule
    & \multicolumn{2}{c}{\textbf{MRA}} & \\
    \cmidrule(lr){2-3}
    \textbf{Model} & Base & 8-task SFT & $\boldsymbol{\Delta\rho}$ \\
    \midrule
    Qwen3-VL 8B & 19.9 & 23.9 & $+0.239$ \\
    Qwen3-VL 4B & 15.9 & 22.4 & $+0.173$ \\
    InternVL3.5 8B & 26.5 & 28.2 & $+0.107$ \\
    \bottomrule
    \end{tabular}}
\end{table}

\newtext{\paragraph{Synchronization without answer options.} The same treatment applies to \taskSync, whose answer is a pair of temporal offsets rather than a category. Standard open-weight checkpoints score 21.8--31.0\% against a 25\% chance rate; the four options do not show how accurately a model can estimate offsets directly. We therefore also ask for the offsets without answer options (Table~\ref{tab:sync_openended}). Because true offsets average 0.70\,s in magnitude, replying ``0 seconds'' everywhere has a mean absolute error (MAE) of 0.70\,s.

The Qwen3-VL-4B and InternVL3.5-8B base checkpoints have MAE of 0.71 and 0.73\,s, close to the constant-zero reference. After fine-tuning, Qwen3-VL-4B reaches MAE 0.48\,s and $\rho = 0.75$, while InternVL3.5-8B reaches MAE 0.37\,s and $\rho = 0.90$.}

\newtext{MAE is in seconds; $\rho$ is the Spearman correlation between predicted and true offsets; $|\text{pred}|$ is the mean magnitude of the predicted offsets.}

\begin{table}[htbp]
  \centering
  \color{black}
  \captionsetup{font+=appendixblackfont,labelfont+=appendixblackfont}
  \caption{\newtext{\textbf{\taskSync\ without answer options.} Models report the two offsets as numbers rather than choosing among four candidate pairs, on 200 items with both offsets scored (400 values).}}
  \label{tab:sync_openended}
  \adjustbox{max width=\linewidth}{
    \begin{tabular}{lcccccc}
    \toprule
    \textbf{Model} & \textbf{MAE (s)} & \textbf{$\boldsymbol{<}$0.25\,s} & \textbf{$\boldsymbol{<}$0.5\,s} & \textbf{Sign} & $\boldsymbol{\rho}$ & $\boldsymbol{|\text{pred}|}$ \\
    \midrule
    \textit{Always answer 0} & 0.70 & 24.5 & --- & --- & --- & 0.00 \\
    \midrule
    Qwen3-VL 4B & 0.71 & 24.2 & 43.0 & 7.8 & $-0.043$ & 0.10 \\
    \quad + 8-task SFT & \textbf{0.48} & 33.0 & 61.5 & 73.0 & $\mathbf{+0.754}$ & 0.60 \\
    InternVL3.5 8B & 0.73 & 21.5 & 38.8 & 47.0 & $-0.006$ & 0.33 \\
    \quad + 8-task SFT & \textbf{0.37} & \textbf{48.5} & \textbf{73.5} & 68.2 & $\mathbf{+0.900}$ & 0.41 \\
    \bottomrule
    \end{tabular}}
\end{table}

\newtext{\subsection{Controls on the Training Gains}
\label{app:training_controls}

Each control below tests one way the reported gains could be an artefact rather than learning: supervision that is not cross-video, answer-format familiarity, a single training seed, text regularities rather than video evidence, multiple testing, and priors installed by fine-tuning.}

\newtext{\subsubsection{Alternative Supervision at a Matched Budget}
\label{app:supervision_arms}

The placebo tests the role of label correctness, but does not establish whether cross-video supervision is necessary: other spatially grounded training data might produce similar transfer. We therefore train a second arm on SAT~\citep{ray2024sat}, a published dataset of multi-frame spatial questions, holding the base, LoRA recipe, 15{,}960 examples, prompt and answer format fixed, with each SAT frame rendered as a one-second clip. It does not reproduce the result (Table~\ref{tab:supervision_arms}): on the two real benchmarks SYNCR improves it moves the other way, by $-8.5$ and $-17.0$ points, and on the constructed ordering sets it loses 5.5 points on both where SYNCR gains 9.0 and 19.0; its one gain is VSI-Bench appearance order ($+7.0$), the target closest to its own content, where SYNCR's $+14.5$ is twice as large. Since SAT is image-based and we render it as clips, this tests its content under our format, so the claim is narrow: matched-budget supervision from a published spatial dataset does not produce these gains.}

\begin{table}[htbp]
  \centering
  \color{black}
  \captionsetup{font+=appendixblackfont,labelfont+=appendixblackfont}
  \caption{\newtext{\textbf{Alternative supervision at a matched budget.} SAT~\citep{ray2024sat} trained on the same base, LoRA recipe, 15{,}960 examples, prompt and answer format. Only the content of the supervision differs.}}
  \label{tab:supervision_arms}
  \adjustbox{max width=\linewidth}{
    \begin{tabular}{lccc}
    \toprule
    \textbf{Benchmark} & \textbf{Base} & \textbf{SAT SFT} & \textbf{8-task SYNCR SFT} \\
    \midrule
    MVU-Eval Temporal Reasoning & 68.0 & 59.5 ($-8.5$) & \textbf{75.5} ($+7.5$) \\
    CrossVid Proc.\ Step Sequencing & 74.5 & 57.5 ($-17.0$) & \textbf{84.5} ($+10.0$) \\
    MVU-Eval Comparison & 68.9 & 65.9 ($-3.0$) & 68.9 ($0.0$) \\
    CrossVid BU & 55.0 & 54.0 ($-1.0$) & 56.0 ($+1.0$) \\
    MVU-Eval KIR & 41.0 & 45.0 ($+4.0$) & 43.5 ($+2.5$) \\
    \midrule
    VSI-Bench appearance order & 54.0 & 61.0 ($+7.0$) & \textbf{68.5} ($+14.5$) \\
    Assembly101 ordering & 35.5 & 30.0 ($-5.5$) & \textbf{44.5} ($+9.0$) \\
    Panoptic ordering & 28.5 & 23.0 ($-5.5$) & \textbf{47.5} ($+19.0$) \\
    \bottomrule
    \end{tabular}}
\end{table}

\newtext{\subsubsection{A Format-Matched Placebo}
\label{app:placebo}

We train a placebo adapter on the same 15{,}960 examples, prompts, videos, option sets, and answer format as the 8-task SFT model, replacing each target with a randomly chosen wrong option.

Training on the same inputs and answer format with incorrect labels does not reproduce the gains from correct supervision. The placebo performs below the base model on every SYNCR task and all five existing real-benchmark tasks reported in Table~\ref{tab:placebo}, with losses of 7.5--35.0 points on the latter. Additional evaluations show declines on VSI-Bench appearance order (54.0\% to 38.0\%) and Assembly101 ordering (35.5\% to 28.0\%), where correctly labeled SFT reaches 68.5\% and 44.5\%, respectively. This control supports the importance of correct supervision, but does not isolate the contribution of answer-format familiarity. App.~\ref{app:realvalidity} separately tests whether ordering gains persist under alternative answer formats.}

\begin{table}[htbp]
  \centering
  \color{black}
  \captionsetup{font+=appendixblackfont,labelfont+=appendixblackfont}
  \caption{\newtext{\textbf{Format-matched placebo control.} Same examples, prompts, videos, option sets and answer format as the 8-task model, with every target replaced by a randomly chosen wrong option.}}
  \label{tab:placebo}
  \adjustbox{max width=\linewidth}{
    \begin{tabular}{lccccccccc|ccccc}
    \toprule
    & \textsc{Sync} & \textsc{Order} & \textsc{ReID} & \textsc{Meas} & \textsc{Num} & \textsc{Kin} & \textsc{Count} & \textsc{Route} & & \textbf{TR} & \textbf{PSS} & \textbf{Comp.} & \textbf{KIR} & \textbf{BU} \\
    \midrule
    Qwen3-VL-8B (base)   & 31.5 & 55.0 & 43.5 & 27.0 & 27.0 & 21.0 & 30.5 & 25.5 & & 68.0 & 74.5 & 68.9 & 41.0 & 55.0 \\
    \quad + placebo SFT  & 23.5 & 18.5 & 26.5 & 25.5 & 19.5 & \phantom{0}5.0 & 28.5 & 24.0 & & \textbf{43.0} & \textbf{39.5} & 49.6 & 33.5 & 41.0 \\
    \quad + 8-task SFT   & 70.0 & 98.0 & 73.0 & 39.5 & 50.0 & 75.0 & 34.5 & 52.5 & & \textbf{75.5} & \textbf{84.5} & 68.9 & 43.5 & 56.0 \\
    \bottomrule
    \end{tabular}}
\end{table}

\newtext{\subsubsection{Seed Variation}
\label{app:seeds}

We repeat the 8-task SFT recipe with three training seeds, changing nothing else (Table~\ref{tab:seeds}). The per-task standard deviation is at most 5.1 points, on \taskMeasure, and below 2.5 on six of the eight tasks; \taskOrder\ returns 98.0\% in all three runs. Average accuracy is 61.6\%, 61.8\% and 64.6\%, a mean of 62.6\%; the main text reports the first seed throughout. Task-level differences smaller than the standard deviations here should not be read as effects, which applies in particular to the single-task versus multi-task comparisons of Sec.~\ref{sec:learning} and the transfer matrix of App.~\ref{app:training_full}.}

\begin{table}[htbp]
  \centering
  \color{black}
  \captionsetup{font+=appendixblackfont,labelfont+=appendixblackfont}
  \caption{\newtext{\textbf{Three seeds of the 8-task SFT recipe.} Accuracy (\%) on the first 200 evaluation items per task, for three runs of the same recipe differing only in training seed.}}
  \label{tab:seeds}
  \adjustbox{max width=\linewidth}{
    \begin{tabular}{lcccccccc|c}
    \toprule
    & \textsc{Sync} & \textsc{Order} & \textsc{ReID} & \textsc{Meas} & \textsc{Num} & \textsc{Kin} & \textsc{Count} & \textsc{Route} & \textbf{Avg.} \\
    \midrule
    Seed 1 (reported) & 70.0 & 98.0 & 73.0 & 39.5 & 50.0 & 75.0 & 34.5 & 52.5 & 61.6 \\
    Seed 2 & 69.5 & 98.0 & 74.5 & 46.0 & 48.0 & 72.0 & 34.5 & 52.0 & 61.8 \\
    Seed 3 & 73.5 & 98.0 & 77.5 & 49.5 & 52.5 & 74.5 & 33.5 & 57.5 & 64.6 \\
    \midrule
    Mean & 71.0 & 98.0 & 75.0 & 45.0 & 50.2 & 73.8 & 34.2 & 54.0 & 62.6 \\
    SD & 2.2 & 0.0 & 2.3 & 5.1 & 2.3 & 1.6 & 0.6 & 3.0 & 1.7 \\
    \bottomrule
    \end{tabular}}
\end{table}

\newtext{\subsubsection{Video-Dependence Contrast}
\label{app:video_attr}

A fine-tuned model can improve with or without using the video evidence. Table~\ref{tab:video_attributable} therefore compares each task's training gain with videos to the gain on the same items with videos removed. We define the gain contrast as the full-video gain minus the no-video gain. It is 27.1 points for the eight-task mean, compared with a 28.9-point full-video gain, and exceeds 23 points on five tasks. For \taskRoute, the contrast is only 10.0 points despite a 27.0-point full-video gain, because the model also improves by 17.0 points without videos. This contrast indicates how much larger the measured gain is when videos are present under this ablation; it does not partition the gain into causal visual and textual components.}

\begin{table}[htbp]
  \centering
  \color{black}
  \captionsetup{font+=appendixblackfont,labelfont+=appendixblackfont}
  \caption{\newtext{\textbf{Video-dependence contrast after 8-task SFT.} Accuracy (\%) with and without videos for the base and 8-task SFT models on matched 200-item sets. The gain contrast is the full-video training gain minus the no-video training gain.}}
  \label{tab:video_attributable}
  \adjustbox{max width=\linewidth}{
    \begin{tabular}{lcccccc}
    \toprule
    & \multicolumn{2}{c}{\textbf{Base}} & \multicolumn{2}{c}{\textbf{8-task SFT}} & & \\
    \cmidrule(lr){2-3}\cmidrule(lr){4-5}
    \textbf{Task} & video & no video & video & no video & \textbf{Gain} & \textbf{Gain contrast} \\
    \midrule
    \textsc{Kin} & 21.0 & 26.5 & 75.0 & 28.0 & $+54.0$ & $\mathbf{+52.5}$ \\
    \textsc{Order} & 55.0 & 28.5 & 98.0 & 26.5 & $+43.0$ & $\mathbf{+45.0}$ \\
    \textsc{Sync} & 31.5 & 27.5 & 70.0 & 26.5 & $+38.5$ & $\mathbf{+39.5}$ \\
    \textsc{Num} & 27.0 & 22.5 & 50.0 & 18.0 & $+23.0$ & $\mathbf{+27.5}$ \\
    \textsc{ReID} & 43.5 & 27.0 & 73.0 & 27.0 & $+29.5$ & $\mathbf{+29.5}$ \\
    \textsc{Route} & 25.5 & 22.5 & 52.5 & 39.5 & $+27.0$ & $+10.0$ \\
    \textsc{Meas} & 27.0 & 24.0 & 39.5 & 26.5 & $+12.5$ & $+10.0$ \\
    \textsc{Count} & 30.5 & 27.0 & 34.5 & 28.5 & $+4.0$ & $+2.5$ \\
    \midrule
    \textbf{Mean} & & & & & $+28.9$ & $\mathbf{+27.1}$ \\
    \bottomrule
    \end{tabular}}
\end{table}

\newtext{\subsubsection{Multiplicity Correction}
\label{app:multiplicity}

The paper reports many paired comparisons, so we group them into the families reported together and give Benjamini--Hochberg adjusted values alongside the raw ones. For the eight-task SFT model against its base, seven of the eight SYNCR tasks are significant raw and all seven survive correction (\taskCount\ is significant under neither). Both real-benchmark transfer results survive: MVU-Eval Temporal Reasoning 68.0\% to 75.5\% ($p_{\mathrm{BH}} = 0.021$) and CrossVid Procedural Step Sequencing 74.5\% to 84.5\% ($p_{\mathrm{BH}} = 0.008$). Among the Thinking checkpoint's eleven real-task comparisons none survives, including CVBench Temporal Reasoning ($p_{\mathrm{BH}} = 0.20$), which we therefore do not treat as a finding. The 8-task RL model's MVU-Eval Comparison gain survives ($p_{\mathrm{BH}} = 0.030$). No conclusion drawn in the main text rests on a test that correction removes.}

\newtext{\subsection{Generalization Beyond the Training Configurations}
\label{app:shift}
\textbf{New structure.} Each set in Table~\ref{tab:shift_generalization}(a) changes one task's structure while retaining its simulator, wording, and option design. \taskOrder\ cuts the benchmark's 200 CLEVRER validation videos into three or five gapped segments instead of four. \taskNumerical\ and \taskKinematic\ compare three validation videos instead of two; \taskKinematic\ options contain each video's fastest object and one video's second-fastest. \taskSync\ retains two videos from the first 200 benchmark items and asks for one offset. For the held-out camera rig (rig~B), we render 200 new Kubric scenes with the simulation settings of App.~\ref{app:kubric} but different cameras. Rig~B moves the three cameras' base positions to $(-3.20, -10.80, 8.60)$, $(10.60, 3.40, 3.10)$ and $(-7.90, 7.60, 10.40)$, widens the per-scene position jitter from uniform $[0, 1)$ to uniform $[-1.5, 1.5)$ on each axis, and draws the focal length per scene from 28, 35 and 42\,mm instead of fixing it at 35\,mm. In the rendered scenes, each camera lies 12.5--23.5 units from its rig-A counterpart, against a mean scene-to-scene deviation of 0.48 units within the released rig. Scene seeds are offset so that no released scene is repeated, and the task is rebuilt with the released generator, option construction and letter balancing. \taskCount\ queries four categories in validation scenes. Each set has 200 items with balanced answer letters, and audited text-only rules stay within 6 points of chance. \taskRoute\ is omitted because training already spans four to nine hops; structural shifts for \taskReID\ and \taskMeasure\ would require new rendering.

\textbf{New simulator.} Table~\ref{tab:shift_generalization}(b) keeps each task's standard structure and wording but draws the videos from an engine that the task never used in training. \taskOrder, trained on CLEVRER, is cut in the same way from single Kubric camera views (5\,s), Habitat walkthroughs (20\,s, gaps of 1.2--2.4\,s) and Habitat route clips. \taskCount, trained on Habitat, asks for the distinct objects of two shapes, or of three material--shape categories, across the three cameras of one Kubric scene; each camera's metadata lists only the objects it sees, so objects are matched across cameras by shape and starting position, and items with objects visible in only one or two frames are dropped. Kubric videos come from the 627 scenes reserved for the Kubric evaluation sets and Habitat videos from the validation scenes, so no training video is reused. We fixed each set before evaluating any model on it, including the later CLEVRER swap of \taskCount, which counts two shapes across three consecutive windows of one video, with totals balanced over 3, 4 and 5 or more. \taskNumerical\ has no swap, because Kubric logs every contact, including resting contacts, so collision counts have no clean ground truth; \taskSync\ and \taskMeasure\ need moving multi-camera scenes, which only Kubric provides. Text-only rules stay within 7 points of chance on every set. These tests hold out task--engine combinations, rather than engines entirely absent from the training mix.}

\newtext{\subsection{Sim-to-Real Transfer}
\label{app:realbench}

This subsection completes the sim-to-real analysis of Sec.~\ref{sec:sim2real_transfer}. Table~\ref{tab:fig5_absolute} reports the item counts and matched base and SFT accuracies behind every cell of Fig.~\ref{fig:sim2real_transfer}. Table~\ref{tab:sim2real_transfer} gives additional RL results on selected existing benchmarks, and Table~\ref{tab:real_syncr} gives paired tests for the constructed Assembly101 and Panoptic ordering sets.

\newtext{In Table~\ref{tab:sim2real_transfer}, $^{*}$ marks gains with exact McNemar $p<0.05$. PSS is converted to four-option multiple choice (chance 25\%), and InternVL uses its native video path with at most 16 frames per video, compared with its own base.}

\begin{table}[!htbp]
    \centering
    \caption{\newtext{\textbf{Sim-to-real transfer.} Accuracy (\%) on MVU-Eval~\citep{peng2025mvu}, CrossVid~\citep{li2026crossvid} (200 items per task) and CVBench~\citep{zhu2025cvbench} (all items). Parentheses show changes from each model's own base.}}
    \label{tab:sim2real_transfer}
    \resizebox{\linewidth}{!}{
    \begin{tabular}{llcccccc}
    \toprule
    & & \multicolumn{3}{c}{\textbf{Qwen3-VL-8B}} & \multicolumn{2}{c}{\textbf{InternVL3.5-8B}} \\
    \cmidrule(lr){3-5}\cmidrule(lr){6-7}
    \textbf{Benchmark} & \textbf{Task} & \textbf{Base} & \textbf{8-task SFT} & \textbf{8-task RL} & \textbf{Base} & \textbf{8-task SFT} \\
    \midrule
    MVU-Eval & Knowledge-Intensive Reasoning & 41.0 & 43.5 ($+2.5$) & 40.0 ($-1.0$) & 38.5 & 42.5 ($+4.0$) \\
    MVU-Eval & \textbf{Temporal Reasoning} & 68.0 & \textbf{75.5} ($\mathbf{+7.5^{*}}$) & 68.5 ($+0.5$) & 60.0 & \textbf{76.5} ($\mathbf{+16.5^{*}}$) \\
    \midrule
    CrossVid & Behavioral Understanding & 55.0 & 56.0 ($+1.0$) & 56.0 ($+1.0$) & 45.0 & \textbf{52.5} ($\mathbf{+7.5^{*}}$) \\
    CrossVid & \textbf{Procedural Step Sequencing (MC)} & 74.5 & \textbf{84.5} ($\mathbf{+10.0^{*}}$) & 75.0 ($+0.5$) & 50.5 & \textbf{78.5} ($\mathbf{+28.0^{*}}$) \\
    \midrule
    CVBench & Key-Action Recognition & 63.6 & \textbf{69.3} ($\mathbf{+5.7}$) & 62.5 ($-1.1$) & 61.4 & 67.0 ($+5.7$) \\
    CVBench & Temporal Reasoning & 50.7 & 53.3 ($+2.7$) & 50.7 ($0.0$) & 38.7 & \textbf{52.0} ($\mathbf{+13.3}$) \\
    CVBench & Difference Caption & 61.5 & 65.4 ($+3.8$) & 63.6 ($+2.1$) & 61.8 & 65.5 ($+3.7$) \\
    \bottomrule
    \end{tabular}}
\end{table}

\begin{table*}[t]
\centering
\caption{\textbf{Absolute accuracy behind Fig.~\ref{fig:sim2real_transfer}.} For every column of the figure: the number of evaluation items, base and 8-task SFT accuracy (\%), and the change in percentage points. Both checkpoints of each model use the same items. For CVBench C.~CR, Qwen uses 52 items and InternVL uses 55, as reflected in the reported accuracies. Changes are computed before rounding the displayed accuracies; paired uncertainty for the constructed ordering sets is reported in Table~\ref{tab:real_syncr}.}
\label{tab:fig5_absolute}
\resizebox{\textwidth}{!}{%
\begin{tabular}{llr*{3}{rrc}}
\toprule
& & & \multicolumn{3}{c}{\textbf{Qwen3-VL-8B}} & \multicolumn{3}{c}{\textbf{Qwen3-VL-4B}} & \multicolumn{3}{c}{\textbf{InternVL3.5-8B}} \\
\cmidrule(lr){4-6} \cmidrule(lr){7-9} \cmidrule(lr){10-12}
\textbf{Source} & \textbf{Task} & $n$ & base & SFT & $\Delta$ & base & SFT & $\Delta$ & base & SFT & $\Delta$ \\
\midrule
\multicolumn{12}{l}{\textit{Constructed on real footage}} \\
Assembly101 & Order & 200 & 35.5 & 44.5 & $+9.0$ & 29.0 & 48.0 & $+19.0$ & 34.0 & 48.0 & $+14.0$ \\
Panoptic & Order & 200 & 28.5 & 47.5 & $+19.0$ & 26.5 & 43.0 & $+16.5$ & 32.0 & 52.5 & $+20.5$ \\
 & ReID & 200 & 11.0 & 18.5 & $+7.5$ & 17.0 & 22.5 & $+5.5$ & 17.0 & 23.5 & $+6.5$ \\
EPFL & Order & 200 & 33.0 & 42.5 & $+9.5$ & 29.5 & 38.5 & $+9.0$ & 30.5 & 45.0 & $+14.5$ \\
 & Sync & 200 & 8.5 & 26.5 & $+18.0$ & 19.5 & 26.0 & $+6.5$ & 30.0 & 25.0 & $-5.0$ \\
nuScenes & Count & 179 & 17.9 & 31.8 & $+14.0$ & 19.6 & 41.3 & $+21.8$ & 35.8 & 31.8 & $-3.9$ \\
 & Kin. & 124 & 29.0 & 41.1 & $+12.1$ & 40.3 & 45.2 & $+4.8$ & 21.0 & 41.9 & $+21.0$ \\
 & ReID & 103 & 41.7 & 65.0 & $+23.3$ & 39.8 & 59.2 & $+19.4$ & 37.9 & 51.5 & $+13.6$ \\
 & Meas & 70 & 58.6 & 61.4 & $+2.9$ & 48.6 & 65.7 & $+17.1$ & 54.3 & 54.3 & $+0.0$ \\
\midrule
\multicolumn{12}{l}{\textit{Existing benchmarks}} \\
MVU-Eval & KIR & 200 & 41.0 & 43.5 & $+2.5$ & 39.5 & 44.0 & $+4.5$ & 38.5 & 42.5 & $+4.0$ \\
 & TR & 200 & 68.0 & 75.5 & $+7.5$ & 58.5 & 73.0 & $+14.5$ & 60.0 & 76.5 & $+16.5$ \\
CrossVid & BU & 200 & 55.0 & 56.0 & $+1.0$ & 59.0 & 58.0 & $-1.0$ & 45.0 & 52.5 & $+7.5$ \\
 & PSS & 200 & 74.5 & 84.5 & $+10.0$ & 58.5 & 73.0 & $+14.5$ & 50.5 & 78.5 & $+28.0$ \\
CVBench & M. KAR & 88 & 63.6 & 69.3 & $+5.7$ & 62.5 & 63.6 & $+1.1$ & 61.4 & 67.0 & $+5.7$ \\
 & M. TR & 75 & 50.7 & 53.3 & $+2.7$ & 40.0 & 50.7 & $+10.7$ & 38.7 & 52.0 & $+13.3$ \\
 & C. CR & 52/55 & 61.5 & 65.4 & $+3.8$ & 48.1 & 55.8 & $+7.7$ & 61.8 & 65.5 & $+3.7$ \\
VSI-Bench & AO & 200 & 54.0 & 68.5 & $+14.5$ & 51.0 & 59.0 & $+8.0$ & 46.0 & 52.0 & $+6.0$ \\
 & RP & 64 & 21.9 & 34.4 & $+12.5$ & 28.1 & 21.9 & $-6.2$ & 20.3 & 34.4 & $+14.1$ \\
\bottomrule
\end{tabular}}
\end{table*}

\subsubsection{Constructed Sets on Real Footage}
\label{app:real_syncr}
\label{app:nuscenes}

\paragraph{SYNCR tasks on real footage.}
We construct two ordering sets from Assembly101~\citep{sener2022assembly101} and CMU Panoptic~\citep{joo2015panoptic} footage using SYNCR's task format. Each contains 200 questions, with balanced answer letters; both were fixed before model evaluation (generation seed 20260917). Ground truth comes from the segment chronology. These sets test transfer while retaining the synthetic tasks' question formats.

\textbf{Ordering.} We sample a 20-second window from each selected source recording, divide it into four segments separated by 1.2--2.4-second gaps, and shuffle the segments. The correct option restores their original chronology, using SYNCR's ordering distractor builder. Assembly101 uses static RGB recordings; Panoptic uses individual HD-camera streams from six sequences, with multiple windows per sequence. Models receive cropped clips without source timestamps, sampled at 2 FPS.

\newtext{\textbf{EPFL ordering and synchronization.} Both sets use the EPFL multi-camera pedestrian videos~\citep{fleuret2007pom}: the laboratory sequences \texttt{4p} and \texttt{6p} (four cameras each) and the campus sequence \texttt{campus4} (three cameras), giving 11 synchronized fixed-camera streams at $360 \times 288$. A recording is kept only if its camera files agree in duration to within 0.1\,s. Neither set uses EPFL's annotations; as for Assembly101 and Panoptic, the answer follows from how the clips are cut. The ordering set cycles through the 11 streams in shuffled order, so each stream contributes 18 or 19 of the 200 items, and each item is built as above, with the 20-second window at least 8\,s from either end of the recording. The synchronization set rotates over the three recordings. Each item assigns three of the recording's cameras, in random order, to Videos~1--3, draws a reference time uniformly between 8\,s after the start and 12\,s before the end, and draws three distinct start offsets of 0--24 frames at 12 FPS (0--2\,s); each video is a 3-second (36-frame) crop beginning at the reference time plus its offset. The correct option gives the start times of Videos~2 and~3 relative to Video~1, and the three distractors come from independent repetitions of the same offset sampling, as in \taskSync\ (App.~\ref{app:kubric}). Each set has 200 items with 50 correct answers per letter (generation seed 20260921).}

\newtext{\textbf{Panoptic re-identification.} This set uses CMU Panoptic's 3D body poses (\texttt{hdPose3d\_stage1\_coco19}), which give each tracked person a persistent identity, together with the HD-camera calibration. Candidate windows start every 4\,s, and a window qualifies only if at least four people are tracked; only the sequence \texttt{160422\_ultimatum1} meets this. For each ordered pair of its HD cameras, people are ranked by the distance from the camera centre to their body centre (the mean of joints with confidence above 0.1, requiring at least five such joints), and people behind the camera are excluded. The question names one person by depth rank in Video~1, drawn from nearest to 4th nearest. That person must be at least 1.15 times farther from the camera than the next-nearer person and at least 1.15 times nearer than the next-farther one, so that the rank identifies a single person, and must be among the four nearest people in Video~2. The question asks for the person's rank in Video~2; the four options are the fixed ranks from nearest to 4th nearest, so the distractors are the other three ranks. From the 739 qualifying candidates in random order, we keep the first 200 that give 50 items per correct rank. Each video is a 3-second clip from one HD camera beginning at the annotated frame, resampled to 12 FPS.}

\newtext{\textbf{nuScenes spatial measurement.} This set uses the six surround cameras of nuScenes (v1.0-mini) and windows of six consecutive keyframes (3\,s at 2 FPS), starting every two keyframes. An object can be named in a camera only if it belongs to one of eight categories (car, pedestrian, truck, bicycle, motorcycle, bus, trailer, construction vehicle), its projected box covers at least 0.4\% of the frame in at least three of the six keyframes, and it is the only visible instance of its category in that camera over the window, so that ``the bus in Video~2'' refers to one object. Each item names four such objects drawn from at least three cameras; the videos are those cameras' clips, and each option reads ``Video~$k$: the $\langle$category$\rangle$''. The question asks which object is farthest from, or nearest to, the recording vehicle. Range is the ground-plane distance from the annotated box centre to the vehicle's ego pose, averaged over the window, so objects in different cameras are compared in one vehicle frame. The correct option is the object at the requested extreme, the other three named objects are the distractors, and an item is kept only if the answer beats the runner-up by at least 6\,m. Because large objects such as buses remain above the visibility threshold at longer range, answer category is confounded with range; we therefore ask both directions and cap each answer category at nine items per direction. We also keep at most two items per window and rotate the correct option through letters A--D, which yields 70 items.}

\newtext{Chance is 25\%; where reported, $p$ is the exact paired McNemar test and the interval is a 95\% bootstrap interval over items (10{,}000 resamples). The ordering sets are also run on the RL models, on InternVL3.5-8B and on Qwen3-VL-4B, each against its own base. Paired results for the unchanged SFT comparisons have intervals excluding zero; the pure-RL arm gains 3.0 points on Assembly101 at $p = 0.21$. For Panoptic Qwen3-VL-8B, the corrected baseline has 57 correct items, versus 95 for SFT and 96 for SFT$\rightarrow$RL. Even under the least favorable possible pairing of these outcomes, the exact McNemar $p$ is below 0.003 for both SFT comparisons. Their bootstrap intervals and the pure-RL paired test require item-level predictions and are left unreported.}

\begin{table}[htbp]
\centering
\caption{\textbf{Ordering on real footage.} Accuracy (\%) on 200 items per set. Chance is 25\%. A dagger denotes a conservative upper bound on the exact paired $p$ over all pairings consistent with the corrected marginal scores; dashes indicate statistics that require item-level predictions.}
\label{tab:real_syncr}
\small
\begin{tabular}{lrrrcr}
\toprule
\textbf{Footage} & \textbf{Base} & \textbf{Model} & $\boldsymbol{\Delta}$ & \textbf{95\% CI} & $\boldsymbol{p}$ \\
\midrule
\multicolumn{6}{l}{\textit{Qwen3-VL-8B, 8-task SFT}} \\
Assembly101 & 35.5 & 44.5 & $+9.0$ & $[+2.5, +16.0]$ & 0.015 \\
Panoptic & 28.5 & 47.5 & $+19.0$ & --- & $<0.003^{\dagger}$ \\
\midrule
\multicolumn{6}{l}{\textit{Qwen3-VL-8B, 8-task RL}} \\
Assembly101 & 35.5 & 38.5 & $+3.0$ & $[-1.0, +7.0]$ & 0.21 \\
Panoptic & 28.5 & 33.5 & $+5.0$ & --- & --- \\
\midrule
\multicolumn{6}{l}{\textit{Qwen3-VL-8B, SFT$\rightarrow$RL}} \\
Assembly101 & 35.5 & 44.5 & $+9.0$ & $[+2.0, +16.0]$ & 0.015 \\
Panoptic & 28.5 & 48.0 & $+19.5$ & --- & $<0.003^{\dagger}$ \\
\midrule
\multicolumn{6}{l}{\textit{InternVL3.5-8B, 8-task SFT}} \\
Assembly101 & 34.0 & 48.0 & $+14.0$ & $[+6.5, +21.5]$ & $9\times10^{-4}$ \\
Panoptic & 32.0 & 52.5 & $+20.5$ & $[+11.5, +29.5]$ & $2\times10^{-5}$ \\
\midrule
\multicolumn{6}{l}{\textit{\newtext{Qwen3-VL-4B, 8-task SFT}}} \\
\newtext{Assembly101} & \newtext{29.0} & \newtext{48.0} & \newtext{$+19.0$} & \newtext{$[+11.5, +26.5]$} & \newtext{$2\times10^{-6}$} \\
\newtext{Panoptic} & \newtext{26.5} & \newtext{43.0} & \newtext{$+16.5$} & \newtext{$[+9.0, +24.0]$} & \newtext{$5\times10^{-5}$} \\
\bottomrule
\end{tabular}
\end{table}

Ordering improves on both sources (Table~\ref{tab:real_syncr}), while synchronization remains near chance. The corresponding ordering gains for InternVL3.5-8B are 14.0 and 20.5 points (Fig.~\ref{fig:sim2real_transfer}). These results distinguish transfer of chronological ordering from synchronization across real camera views.

\newtext{\paragraph{Cross-camera counting on driving footage.}
The ordering sets above preserve SYNCR's question format using segments from one recording.
To examine simultaneous camera views, we
build a counting set from nuScenes~\citep{caesar2020nuscenes}, whose six synchronized
surround cameras carry 3D boxes with an instance token identifying the same physical object
in every camera that sees it. Each item presents six clips, one per camera, built from six
consecutive keyframes of one scene at 2~FPS, and asks how many distinct objects of a named
category appear across all six views. Visibility is decided by the nuScenes devkit's own
projection rather than by a rule of ours. The set has 179 items over six categories with
true counts from 2 to 12; in 79.9\% of them at least one object is visible in two or more
cameras.

We score each model on the same items with all six cameras and with one randomly selected
camera (Table~\ref{tab:nuscenes_cameras}). The accuracy difference measures the benefit of
supplying additional views under this protocol. It does not identify whether the model
matches repeated sightings across cameras or benefits from seeing objects absent from the
selected single view.

All three base checkpoints shown in the table score higher with six cameras. For the first
Qwen3-VL-8B SFT seed, the six-versus-one gap rises from $+7.8$ to $+17.3$ points; for
Qwen3-VL-4B, it rises from $+7.8$ to $+18.4$ points. The increases over their respective
bases are $+9.5$ points ($p=0.011$) and $+10.6$ points ($p=0.039$) by a paired bootstrap over
items. The increase is not established for two further Qwen3-VL-8B seeds or for
InternVL3.5-8B. Absolute accuracy remains low, and these results do not establish reliable
transfer of cross-camera counting.} \newtext{Two further cross-camera sets use the same footage: \taskKinematic\ asks which of four objects across the views is moving fastest, with ground truth from the annotated velocities, and \taskReID\ asks in which window of a second camera an object first reappears, the object named by its range from the vehicle so that the reference is unambiguous.}

\newtext{In Table~\ref{tab:nuscenes_cameras}, $p$ and the interval are as in App.~\ref{app:real_syncr}, with the test taken between the two conditions. Every checkpoint shown answers better with six cameras than with one. Whether fine-tuning enlarges the gap is not established across runs: the increase is supported for the first Qwen3-VL-8B seed and at 4B, but not for two further Qwen3-VL-8B seeds (omitted from the table) or for InternVL3.5-8B.}

\begin{table}[htbp]
  \centering
  \color{black}
  \captionsetup{font+=appendixblackfont,labelfont+=appendixblackfont}
  \caption{\newtext{\textbf{Cross-camera counting on nuScenes driving footage.} Accuracy (\%) on the same 179 items given all six synchronized cameras and given a single randomly chosen camera. Chance is 25\%.}}
  \label{tab:nuscenes_cameras}
  \adjustbox{max width=\linewidth}{
    \begin{tabular}{lrrrcr}
    \toprule
    \textbf{Model} & \textbf{One camera} & \textbf{Six cameras} & \textbf{Gap} & \textbf{95\% CI} & $\boldsymbol{p}$ \\
    \midrule
    Qwen3-VL-8B base & 10.1 & 17.9 & $+7.8$ & $[+2.8, +12.8]$ & $4\times10^{-3}$ \\
    \quad + 8-task SFT & 14.5 & 31.8 & $+17.3$ & $[+10.6, +24.0]$ & $2\times10^{-6}$ \\
    \midrule
    Qwen3-VL-4B base & 11.7 & 19.6 & $+7.8$ & $[+1.7, +14.0]$ & $0.029$ \\
    \quad + 8-task SFT & 22.9 & 41.3 & $+18.4$ & $[+10.6, +26.3]$ & $2\times10^{-5}$ \\
    \midrule
    InternVL3.5-8B base & 22.3 & 35.8 & $+13.4$ & $[+6.7, +20.1]$ & $3\times10^{-4}$ \\
    \quad + 8-task SFT & 21.2 & 31.8 & $+10.6$ & $[+4.5, +17.3]$ & $1\times10^{-3}$ \\
    \bottomrule
    \end{tabular}}
\end{table}

\subsubsection{Existing Multi-Video Benchmarks}
\label{app:realvalidity}

\newtext{\paragraph{PSS in answer formats SYNCR never produces.}

\newtext{Because PSS shares SYNCR's ``Video~$i \rightarrow$ Video~$j$'' ordering-answer format, a gain there could reflect format familiarity rather than transfer, so we re-ran the same items, clips and correct orders in two answer formats SYNCR never produces: \emph{letters}, where the clips are relabelled A--E and the answer is a letter string (``C, E, D, B, A''), and \emph{prose}, where the answer is a sentence (``the third clip, then the fifth clip, \ldots''). On 200 items, 8-task SFT improves Qwen3-VL-8B by 19.0 points in the letters format (57.0\% to 76.0\%, $p < 10^{-5}$) and 8.0 in prose (77.5\% to 85.5\%, $p = 0.011$), against 10.0 in SYNCR's own format (74.5\% to 84.5\%, $p = 0.0012$); InternVL3.5-8B replicates it more strongly still, gaining 30.5 points in letters and 30.0 in prose with its 8-task model (Table~\ref{tab:pss_format}).}

\begin{table}[htbp]
  \centering
  \color{black}
  \captionsetup{font+=appendixblackfont,labelfont+=appendixblackfont}
  \caption{\newtext{\textbf{CrossVid PSS in answer formats SYNCR never produces.} The same items, clips and correct orders in a letter-string and a sentence format; gains over each base in parentheses.}}
  \label{tab:pss_format}
  \adjustbox{max width=\linewidth}{
    \begin{tabular}{lccc}
    \toprule
    \textbf{Model} & \textbf{SYNCR's format} & \textbf{Letters} & \textbf{Prose} \\
    \midrule
    Qwen3-VL-8B & 74.5 & 57.0 & 77.5 \\
    \quad + 8-task SFT & 84.5 ($+10.0$) & 76.0 ($+19.0$) & 85.5 ($+8.0$) \\
    \midrule
    InternVL3.5-8B & 50.5 & 30.5 & 42.5 \\
    \quad + 4-task SFT & 64.5 ($+14.0$) & 44.0 ($+13.5$) & 53.0 ($+10.5$) \\
    \quad + 8-task SFT & 78.5 ($+28.0$) & 61.0 ($\mathbf{+30.5}$) & 72.5 ($\mathbf{+30.0}$) \\
    \bottomrule
    \end{tabular}}
\end{table}

\newtext{ The gain is largest in the format the training data never contains, which is the opposite of what format familiarity predicts, so we treat PSS as measuring transferred ordering rather than answer-template reuse.}

}

\newtext{\paragraph{RL from the base model.} Qwen3-VL 8-task RL changes MVU-Eval Temporal Reasoning and CrossVid sequencing by 0.5 points each and improves constructed real-video ordering by 3.0--5.0 points, well below the corresponding SFT gains. The exception is VSI-Bench appearance order, where RL gains 13.5 points and SFT$\rightarrow$RL 14.0, as much as SFT: on that single-video benchmark the supervised stage is not what separates the models.}

\paragraph{Temporal transfer after RL on SFT.}
Applying GRPO to the 8-task SFT model leaves both ordering gains in place: MVU-Eval Temporal Reasoning is 76.5\% (+8.5 over the base, exact McNemar $p = 0.0009$) and CrossVid PSS 84.0\% (+9.5, $p = 0.0019$). Against its own SFT starting point, no task changes significantly in either direction (Temporal Reasoning +1.0, $p = 0.63$; PSS $-0.5$, $p = 1.00$), and on PSS it is 9.0 points above RL from the base model ($p = 0.003$). CVBench Temporal Reasoning rises by 8.0 points ($p = 0.18$), the largest CVBench change for this model, without reaching significance on 75 items.

\paragraph{Comparison with reasoning-specialized post-training.}
Qwen3-VL-8B-Thinking, evaluated zero-shot, improves CVBench Temporal Reasoning from 50.7\% to 66.7\% (raw $p = 0.029$), the largest difference from Instruct among the eleven tasks. \newtext{It does not survive multiplicity correction within that family ($p_{\mathrm{BH}} = 0.20$; App.~\ref{app:multiplicity}), so we do not treat it as a finding.} Its CrossVid PSS is 81.5\% (+7.0, $p = 0.059$), compared with 84.5\% for 8-task SFT. Its MVU-Eval Temporal Reasoning accuracy is 73.0\% ($p = 0.093$ versus Instruct). Both the Thinking checkpoint and SYNCR SFT show higher accuracy on real temporal tasks, consistent with the ordering improvement observed for Thinking on SYNCR (Sec.~\ref{sec:main_results}).

\paragraph{Larger temporal-transfer gains under InternVL's eight-task mix.}
InternVL3.5-8B benefits from both training mixes, with 8-task SFT producing approximately twice the ordering gains of 4-task SFT (+28.0 versus +14.0 on PSS and +16.5 versus +8.0 on Temporal Reasoning). The 8-task model, which adds four tasks including \taskOrder, significantly outperforms 4-task on both ordering benchmarks (PSS $p = 0.0004$, Temporal Reasoning $p = 0.0076$) and on CrossVid BU ($p = 0.0044$). The 4-task mix, trained on \taskSync, \taskReID, \taskNumerical\ and \taskCount, itself improves PSS by 14.0 points ($p = 2.3 \times 10^{-4}$) and Temporal Reasoning by 8.0 ($p = 0.0052$); on the other nine tasks it does not differ significantly from either the base or 8-task model. The eight-task mix also uses more examples, so the comparison does not isolate which added task or the extra training exposure accounts for the larger gains.
}

\end{document}